\documentclass[runningheads]{llncs}
\usepackage{graphicx}

\usepackage{tikz}
\usepackage{comment}
\usepackage{adjustbox}
\usepackage{color}
\usepackage{makecell}
\usepackage{booktabs}
\usepackage{multirow}
\usepackage{dirtytalk}
\usepackage{dsfont}
\usepackage{bbm}
\usepackage{graphics}
\usepackage{mwe}
\usepackage{hyperref}

\usepackage{amsmath,mathtools}
\usepackage{amssymb}
\usepackage{placeins}

\usepackage{calrsfs}
\usepackage[mathscr]{eucal}
\usepackage{comment}
\usepackage{eqnarray}
\usepackage{adjustbox}
\usepackage{makecell}
\usepackage{tabularx}
\usepackage[font=footnotesize]{caption}
\usepackage[caption=false]{subfig}
\usepackage{xcolor}
\usepackage{calligra}
\usepackage{mathtools,nccmath,tabstackengine}
\usepackage{array}
\usepackage{multirow}
\hypersetup{
     colorlinks=true,
     linkcolor=black,
     filecolor=black,
     citecolor = black,      
     urlcolor=blue,
     }

\usepackage{physics}

\newcommand{\R}{\mathbb{R}}
\newcommand\tab[1][0.5cm]{\hspace*{#1}}

\begin{document}
\pagestyle{headings}
\mainmatter
\def\ECCVSubNumber{10}  

\title{How to track your dragon:\\
A Multi-Attentional Framework for real-time RGB-D 6-DOF Object Pose Tracking}

\titlerunning{How to track your dragon}

\author{Isidoros Marougkas\inst{1}
\and
Petros Koutras\inst{1},
\and
Nikos Kardaris\inst{1} \and 
Georgios Retsinas\inst{1}
\and \\
Georgia Chalvatzaki\inst{2}
\and \\
Petros Maragos\inst{1}
}

\authorrunning{I.Marougkas et al.}

\institute{$^1$School of E.C.E., National Technical University of Athens, 15773, Athens, Greece\\  
	$^2$Department of Computer Science TU Darmstadt, 64289, Darmstadt, Germany\\
	\small{Email: \{ismarougkas, nick.kardaris\}@gmail.com, \{pkoutras, maragos\}@cs.ntua.gr}, gretsinas@central.ntua.gr, georgia@robot-learning.de}

\maketitle
\begin{abstract}
We present a novel multi-attentional convolutional architecture to tackle the problem of real-time RGB-D 6D object pose tracking of single, known objects. Such a problem poses multiple challenges originating both from the objects' nature and their interaction with their environment, which previous approaches have failed to fully address. The proposed framework encapsulates methods for background clutter and occlusion handling by integrating multiple parallel soft spatial attention modules into a multitask Convolutional Neural Network (CNN) architecture. Moreover, we consider the special geometrical properties of both the object's 3D model and the pose space, and we use a more sophisticated approach for data augmentation during training. The provided experimental results confirm the effectiveness of the proposed multi-attentional architecture, as it improves the State-of-the-Art (SoA) tracking performance by an average score of 34.03 $\%$ for translation and 40.01 $\%$ for rotation, when tested on the most complete dataset designed, up to date, for the problem of RGB-D object tracking. Code will be available in: \href{https://github.com/ismarou/How_to_track_your_Dragon}{https://github.com/ismarou/How\_to\_track\_your\_Dragon} 
\end{abstract}

\section{Introduction}

Robust, accurate and fast object pose estimation and tracking, i.e. estimation of the object's 3D position and orientation, has been a matter of intense research for many years. The applications of such an estimation problem can be found in Robotics, Autonomous Navigation, Augmented Reality, etc. Although the Computer Vision community has consistently studied the problem of object pose estimation and tracking for decades, the recent spread of  affordable and reliable RGB-D sensors like Kinect, along with advances in Deep Learning (DL) and especially the use of CNNs as the new SoA image feature extractors, led to a new era of research and a re-examination of several problems, with central aim the generalization over different tasks. CNNs have achieved ground-breaking results in 2D problems like object classification,
object detection and segmentation. Thus, it has been tempting to the research community to increasingly use them in the more challenging 3D tasks, renouncing traditional algorithms. 

The innate challenges of object pose estimation from RGB-D streams include background clutter, occlusions (both static, from other objects present in the scene, and dynamic, due to possible interactions with a human user), illumination variation, sensor noise, image blurring (due to fast movement) and appearance changes as the object viewpoint alters. 
Moreover, one should account for the pose ambiguity, which is a direct consequence of the object's own geometry, in possible symmetries, 
the challenges of proper parameter representation of rotations and the inevitable difficulties that an effort of forging a model faces, when extracting information about the 3D scene geometry from 2D-projected images.

\begin{figure}[t]
    \centering
   \includegraphics[width=.1\linewidth]{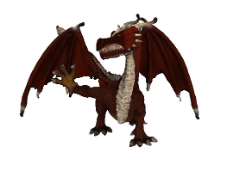}
    \includegraphics[width=.25\linewidth,height=0.2\textwidth]{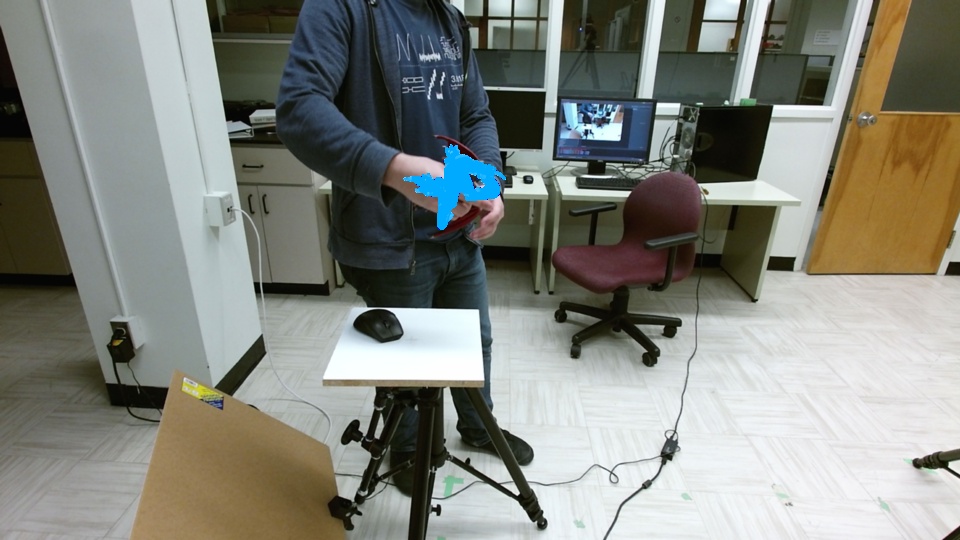} 
    \includegraphics[width=.25\linewidth,height=0.2\textwidth]{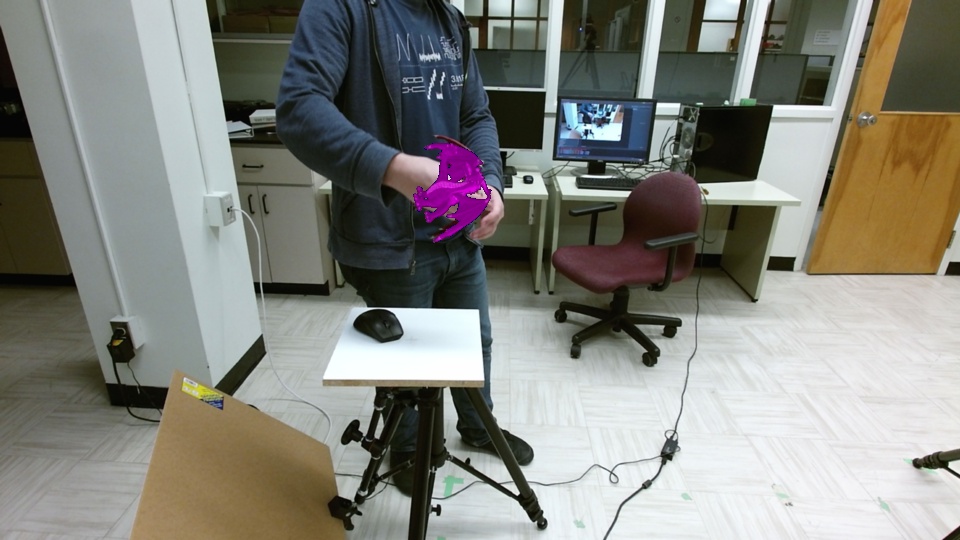} 
    \begin{minipage}[b][0.122\textheight][s]{0.15\textwidth}
  \centering
  \includegraphics[width=.5\textwidth]{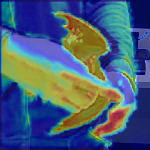}\label{subcap:Foregr}
  \vfill
  \includegraphics[width=.5\textwidth]{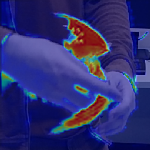}\label{subcap:Occl}
\end{minipage}
    \\
    \includegraphics[width=.27\linewidth]{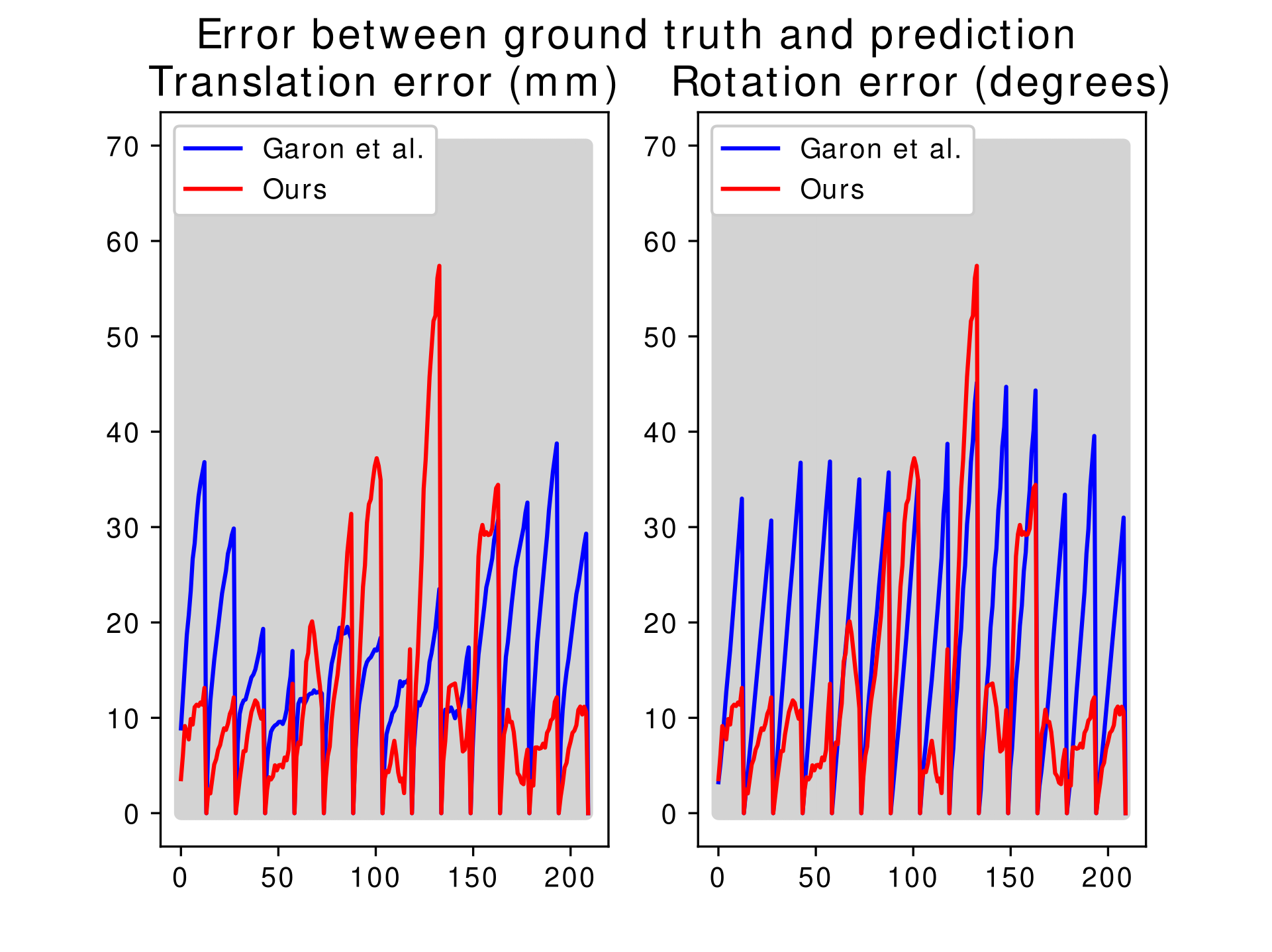}
    \includegraphics[width=.27\linewidth]{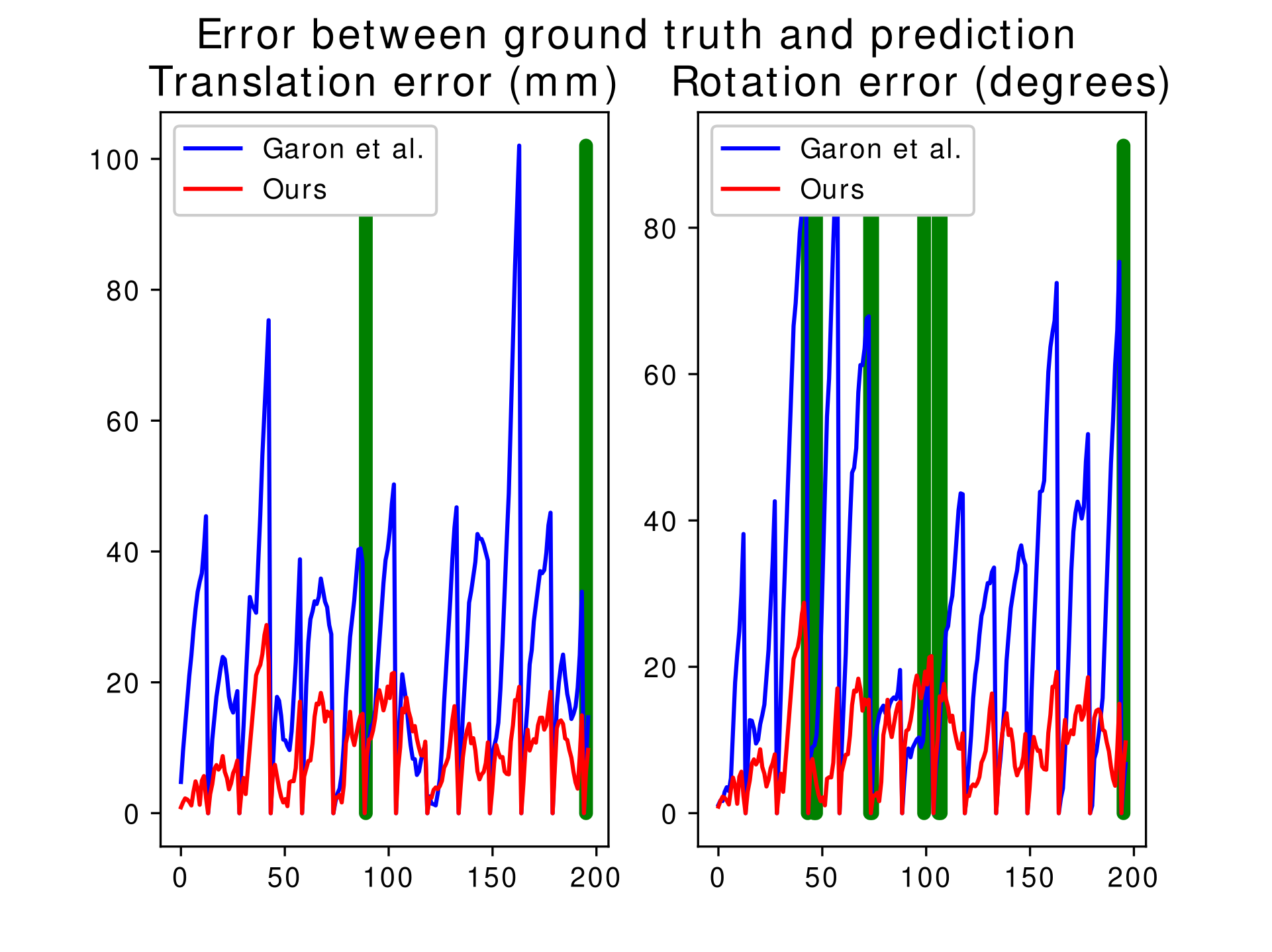}
    \includegraphics[width=.27\linewidth]{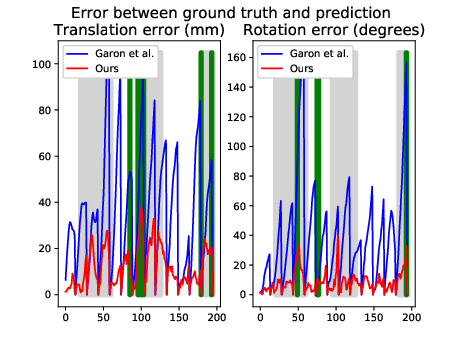}
\vspace{-0.3cm}
\caption{\footnotesize{\textbf{Top row:}\textit{(Left to right)} The "Dragon" model, an estimated pose in the "Hard Interaction" scenario for each of the SoA \cite{Garon_2018} \textit{(light blue)} and our \textit{(pink)} approaches  and an example frame pair of Foreground extraction \textit{(up)} and Occlusion handling \textit{(down)}  attention maps which are learned by minimizing the two auxiliary binary cross entropy losses. The following tradeoff occurs: as the occlusion increases, foreground attention, which focuses on the moving parts of the scene (i.e. the hand and the object), gets blurrier, while occlusion attention gets sharper and shifts focus from the object center to its body parts. \textbf{Bottom row:}\textit{(Left to right)} Translational and Rotational error plots of the SoA \cite{Garon_2018} \textit{(blue)} and \textit{(our)} approaches, for the ``$75\%$ Vert. Occlusion'', the ``Rotation Only'' and the
``Hard Interaction'' scenario, respectively. Grey regions stand for intervals of high occlusion and green ones for rapid movement.}}
\label{subfig:attentions}
\vspace{-0.6cm}
\end{figure}

In this paper, we build upon previous works \cite{Garon_2017,Garon_2018}, in order to face a series of those challenges that have not been fully resolved, so far. Thus, our main contributions are:
\begin{itemize}
     \item 
An explicit background clutter and occlusion handling mechanism that leverages spatial attentions and provides an intuitive understanding of the tracker's region of interest at each frame, while boosting its performance. To the best of our knowledge, this is the first such strategy, that explicitly handles these two challenges, is incorporated into a CNN-based architecture, while achieving real-time performance. Supervision for this mechanism is extracted by fully exploiting the synthetic nature of our training data. 

   \item
The use of a novel multi-task pose tracking loss function, that respects the geometry of both the object's 3D model and the pose space and boosts the tracking performance by optimizing auxiliary tasks along with the principal one.

    \item
 SoA real-time performance in the hardest scenarios of the benchmark dataset \cite{Garon_2018}, while achieving lower translation and rotation errors by an average of $34.03\%$ for translation and $40.01\%$ for rotation.
\end{itemize}
Accordingly, we provide the necessary methodological design details and experimental results that justify the importance of the proposed method in the challenging object pose tracking problem. 

\section{Related Work}

Previous works attempt to tackle the problem using DL, focusing on two different directions: per-frame pose estimation (or, else, ``tracking by detection'') and temporal tracking.
\vspace{-10pt}
\subsubsection{Tracking-by-Detection}
The first family of proposed approaches in literature processes each video frame separately, without any feedback from the estimation of the previous timeframe.  In \cite{Xiang_2017}, Xiang et al. constructed a CNN architecture that estimates binary object masks and then predicts the object class and its translation and rotation separately, while in  \cite{Kehl_2017b}, Kehl et al. extended the Single Shot Detection (SSD) framework \cite{Liu_2016} for 2D Object detection by performing discrete viewpoint classification for known objects. Finally, they refine their initial estimations via ICP \cite{Segal_2009} iterations. In \cite{Zakharov_2019}, Zakharov et al. proposed a CNN framework that uses RGB images for pixel-wise object semantic segmentation in a mask-level. Following this, UV texture maps are estimated to extract dense correspondences between 2D images and 3D object models minimizing cross entropy losses. Those correspondences are used for pose estimation via P'n'P \cite{Lepetit_2009}. This estimation is, ultimately, inserted as a prior to a refinement CNN that outputs the final pose prediction. In PVNet \cite{Peng_2019}, Peng et al. perform per-pixel voting-based regression to match 3D object coordinates with predefined keypoints inside the object surface, in order to handle occlusions. In \cite{Pavlakos_2017}, Pavlakos et al. extract semantic keypoints in single RGB images with a CNN and incorporate them into a deformable shape model. In \cite{Wohlhart_2015}, Wohlhart et al. employed a supervised contrastive convolutional framework to disentangle descriptors of different object instances and impose proportional distances to different poses of the same object. In \cite{Sundermeyer_2018}, Sundermeyer et al. built a self-supervised Augmented Auto-Encoder that predicts 3D rotations only from synthetic data. In \cite{Park_2019}, Park et al. trained an adversiarially guided Encoder-Decoder to predict pixel-wise coordinates in a given image and then fed them to a P'n'P algorithm. More recently, iPose \cite{Jafari_2019} is one of the attempts the philosophy of which is the closest to ours. Its authors segment binary masks with a pretrained MaskRCNN \cite{He_2017} to extract background clutter and occluders and they map 2D pixels to dense 3D object coordinates, which, in turn, are used as input to a P'n'P geometric optimization. Our attention modules  have the same effect, but are computationally cheaper than MaskRCNN, as they relax the requirement for hard segmentation. 

\vspace{-0.4cm}
\subsubsection{Temporal Tracking}
The second category under study is temporal tracking, where  feedback is utilized, to allow for skipping steps without prior knowledge of the previous pose. Garon et al. \cite{Garon_2017,Garon_2018}, formulated the tracking problem exclusively as a learning one, by generating two streams of synthetic RGB-D frame pairs from independent viewpoints and regressing the pose using a CNN. Li et al.\cite{Li_2018} initialized a similar CNN architecture using a FlowNet2 \cite{Ilg_2017} backbone and fused its two streams by subtraction. In DeepTAM \cite{Zhou_2018}, the training was performed with an Optical flow-based regularization term and the production of multiple heterogenous pose hypotheses was encouraged. Those hypotheses were bootstrapped in the final layer. Last but not least, Deng et al.\cite{Deng_2019} extended the framework of \cite{Sundermeyer_2018} by combining it with a Rao-Blackwellized Particle Filter \cite{Doucet_2013}. In brief, they randomly sampled  2D bounding boxes to crop RGB images, infer 3D translation probabilities from their dimensions and search for the closest saved rotated sample provided by the Autoencoder of \cite{Sundermeyer_2018}. Then, the particles were weighted according to the orientation probabilities and were prepared for the next sampling iteration.    
\begin{figure*}[t]
    \centering
    \includegraphics[width=\textwidth]{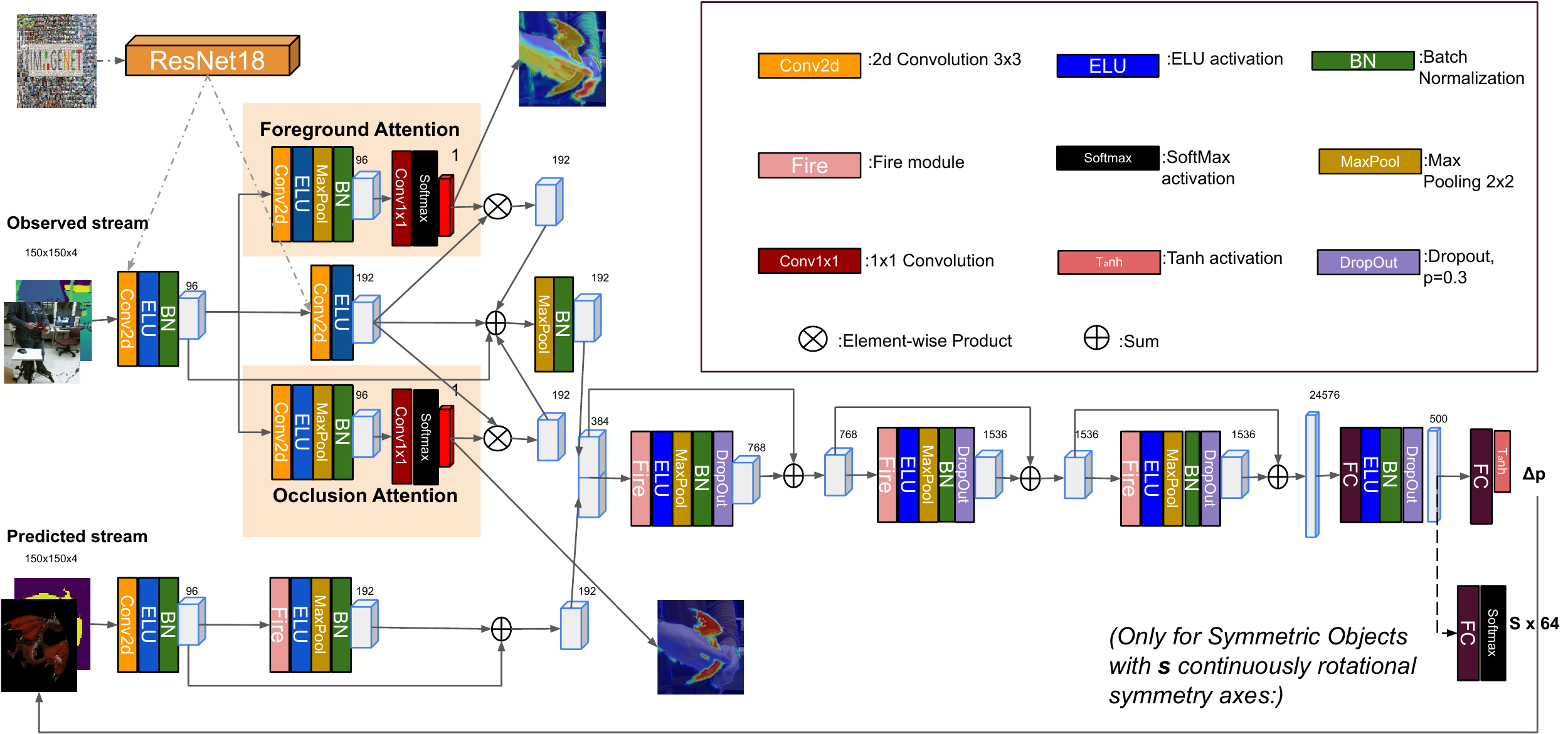}
    \caption{Overview of the proposed CNN architecture for object pose tracking.}
    \label{fig:Ours}
    \vspace{-0.5cm}
\end{figure*}

\vspace{-0.3cm}
\section{Methodology}
\label{sec:methodology}
\subsection{Problem Formulation}\label{subsec:problem_formulation}
Our problem consists in estimating the object pose $\mathbb{P}$, which is usually described as a rigid 3D transformation w.r.t. a fixed coordinate frame, namely an element of the Special Euclidean Lie group in 3D: $SE(3)$. It can be disentangled into two components; a rotation matrix R, which is an element of the Lie Group $\textnormal{SO(3)}$ and a translation vector $\mathbf{t} \in \R^{3}$. 
However, Br\'egier et al. \cite{Bregier_2017} proposed a broader definition for the object pose, which can be considered as a family of rigid transformations, accounting for the ambiguity caused by possible rotational symmetry, noted as  G $\in$ $\textsc{SO(3)}$. We leverage this augmented mathematical definition for introducing a relaxation to the pose space $\mathcal{C}$ definition: 
\vspace{-.2cm}
\begin{equation}
    \resizebox{0.7\linewidth}{!}{ 
    $\mathcal{C} = \Biggl\{ \mathbb{P} \quad | \quad \mathbb{P} = \left[ \begin{array}{c|c} R\cdot G & \mathbf{t} \\ \hline \mathbf{0}^{T} & 1 \end{array} \right],
    \mathbf{t} \in \R^{3}, R \in \textnormal{SO(3)}, G \in \textnormal{SO(3)} \Biggr\}.$}
\end{equation}\\[-0.4cm]
For example, as stated in \cite{Bregier_2017}, the description of the pose of an object with spherical symmetry requires just 3 numbers: ($t_{x,y,z}$), as G can be any instance of $\textnormal{SO(3)}$ with the imprinted shape of the object remaining the same. Obviously, for asymmetrical objects, G=$\mathbb{I}_{3}$.

\vspace{-10.0pt}
\subsection{Architecture Description}\label{subsec:architecture_description}
The proposed architecture is depicted in Fig.\ref{fig:Ours}. 
Our CNN inputs two RGB-D frames of size  $150 \times 150$: I(t),$\hat{{\textnormal{I}}}$(t) (with I(t) being the ``Observed'' and $\hat{\textnormal{I}}$(t) the ``Predicted" one) and regresses an output pose representation $\Delta \mathbf{p} \in \R^{9}$, with 3 parameters for translation ($\hat{t}_{x,y,z} \in [-1,1]$) and 6 for rotation.
The first two layers of the ``Observed'' stream are initialized with the weights of a ResNet18\cite{He_2016}, pretrained on Imagenet \cite{Deng_2009}, to narrow down the real-synthetic domain adaptation gap, as proposed in \cite{Hinterstoisser_2018}. Since Imagenet contains only RGB images, we initialize the weights of the Depth input modality with the average of the weights corresponding to each of the three RGB channels. Contrary to \cite{Hinterstoisser_2018}, we find beneficial not to freeze those two layers during training. The reason is that we aim to track the pose of the single objects we train on and not to generalize to unseen ones. So, overfitting to that object's features helps the tracker to focus only on distinguishing the pose change. The weights that correspond to the Depth stream are, of course, not frozen in either case.  To the output of the second ``Observed'' layer, we apply spatial attention for foreground extraction and occlusion handling and we add their corresponding output feature maps with the one of the second layer, along with a Residual connection \cite{He_2015} from the first layer. As a next step, we fuse the two streams by concatenating their feature maps and pass this concatenated output through three sequential Fire modules \cite{Iandola_2016}, all connected with residual connections \cite{He_2016}.

\smallskip
\noindent\textit{\textbf{{Background and Occlusion Handling:}}}
After our first ``Observed'' Fire layer, our model generates an attention weight map by using a Fire layer dedicated to occlusion handling and foreground extraction, respectively, followed by a $1 \times 1$ convolution that squeezes the feature map channels to a single one (and normalized by softmax). Our goal is to distil the soft foreground and occlusion segmentation masks from the hard binary ground-truth ones (that we keep from augmenting the object-centric image with random backgrounds and occluders) in order to have their estimations available during the tracker's inference. To this end, we add the two corresponding binary cross entropy losses to our overall loss function. We argue our design choice of using two attention modules, as after experimentation, we found that assigning a clear target to each of the two modules is more beneficial, rather than relying on a single attention layer to resolve both challenges (see Sect.\ref{subsubsec:attention_modules_hierarchy}), an observation also reported in \cite{Jafari_2019}.

\smallskip
\hspace{-20pt} 
\noindent
\textit{\textbf{Rotation representation:}} From a mathematical standpoint, immediate regression of pose parameters \cite{Garon_2018} with an Euclidean loss is suboptimal: 
while the translation component belongs to the Euclidean space, the rotation component lies on a non-linear manifold of $\textnormal{SO(3)}$. Thus, it is straightforward to model the rotation loss using a Geodesic metric \cite{Huynh_2009,Hartley_2013} on $\textnormal{SO(3)}$, i.e. the length, in radians, of the minimal path that connects two of its elements: $\Delta \hat{R},\Delta R$:

\vspace{-.2cm}
\begin{equation}
    {\small L_{Rot}(\Delta \hat{R}, \Delta R_{GT}) = d^{(Geod)}_{Rot}(\Delta \hat{R}, \Delta R_{GT}) =\arccos \Big( \frac{\Tr \big( \Delta \hat{R}^{T} \cdot \Delta R_{GT} \big)-1}{2} \Big).}
\end{equation}\\[-0.4cm]
In order to minimize the rotation errors due to ambiguities caused by the parameterization choice, 
we employ the  6D continuous rotation representation that was introduced in \cite{Zhou_2019}: $\Delta \mathbf{r} = (\Delta \mathbf{r_x}^{T}, \Delta \mathbf{r_y}^{T})$, where $\Delta \mathbf{r_{x/y}} \in \R^{3}$. Given $\Delta \mathbf{r}$, the matrix $ \Delta R = (\Delta \mathbf{R_x},\Delta  \mathbf{R_y},\Delta \mathbf{R_z})^T$ is obtained by:
\vspace{-0.2cm}
\footnotesize
\begin{equation}\label{eqn:6D_Cont_Rot}
\begin{split}
    &\Delta \mathbf{R_x} = N(\Delta \mathbf{r_x}) \\
    &\Delta \mathbf{R_y} = N[\Delta \mathbf{r_{y}}-(\Delta \mathbf{R^{T}_{x}} \cdot \mathbf{r_{y}) \cdot \Delta \mathbf{R_{x}}})] \\
    &\Delta \mathbf{R_z} = \Delta  \mathbf{R_x} \times \Delta \mathbf{R_y}
    \end{split}
\end{equation}\\[-0.4cm]
\normalsize
where $\Delta \mathbf{R_{x/y/z}} \in \R^{3}$, $N(\cdot)=\frac{(\cdot)}{\norm{(\cdot)}}$ is the normalization function. Furthermore, as it has already been discussed in \cite{Bregier_2017}, each 3D rotation angle has a different visual imprint regarding each rotation axis. So, we multiply both rotation matrices with an approximately diagonal Inertial Tensor $\Lambda$, calculated on the object model's weighted surface and with respect to its center mass, in order to assign a different weight to each rotational component. We note here that since we want that matrix product to still lie in SO(3), we perform a Gramm-Schmidt orthonormalization on the Inertial Tensor $\Lambda$ before right-multiplying it with each rotation matrix. Finally, we weigh the translation and rotation losses using a pair of learnable weights $\mathbf{v}=[v_{1},v_{2}]^{T}$ that are trained along with the rest of the network's parameters using Gradient Descent-based optimization, as proposed in \cite{Kendall_2017}.

\medskip
\noindent
\textit{\textbf{Symmetric Object Handling}}
In the special case of symmetric objects, we disentangle the ambiguities inserted due to this property from the core of rotation estimation. We classify such ambiguities to two distinctive categories: a continuous set of rotational ambiguities (the unweaving of which we incorporate into our loss function) and a discrete set of reflective ambiguities that appear due to our rotation  representation choice and are handled heuristically.  

Normally, the presence of the RGB input cue would break any symmetry ambiguities as their origin is its Depth counterpart that depicts the object's 3D shape. However, our preliminary experiments have shown that this is only partially true since the tracker places more emphasis to the Depth cue, in general, in its effort to estimate the object's pose. This inclination keeps the ambiguities present during inference and incurs the need to explicitly model this disentanglement in the loss function formulation.

\medskip
\noindent
\textit{\textbf{Symmetric Object Handling: Discrete Reflective Symmetry}}

By replacing the 3D Euler rotational parameter regression of Garon et al.\cite{Garon_2018}, with its 6D continuous counterpart, we face the extra problem of being unable to constrain the network's rotational output. This has a severely negative effect to objects with reflective symmetry, as there are configurations in which the tracker predicts unacceptable values for one or more rotational components. For example, in Fig. \ref{fig:reflective} we observe that the ``Cookie Jar'' symmetric object has been turned upside down with respect to both the prediction of the previous timeframe and the ground truth pose . As a result, we end up with adding significant extra errors during the mean rotational error calculation over the total motion length, since that discrepancy is not limited to a single frame, but is accumulated as we proceed to the following frames, due to the temporal nature of the tracker, until it is reset. In order to handle this challenge, we employ the following heuristic algorithm: for each Euler rotational component we calculate the angular distance $ d^{(o)}_{A}(\hat{r}_{i}(t),\hat{r}_{i}(t-1))$  (in degrees), between the current and the previous timestep.
If a $d^{(o)}_{A}$ exceeds a certain threshold (here, it is set to $100^{o}$), then the value of this particular Euler angle is set to $\hat{r_{i}}(t-1)$. Then, we may choose to perform a second (or more) iterative forward CNN pass(es), if the time constraints of our broader application allow so.

\begin{figure}[t]
	\centering
	\begin{minipage}[t]{0.327\textwidth}
		\includegraphics[width=0.96\textwidth]{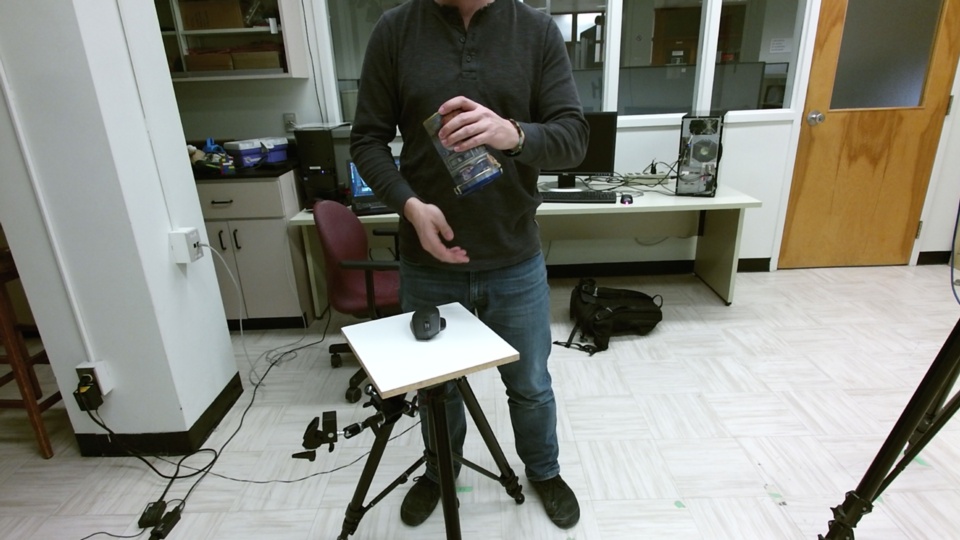}\captionsetup{singlelinecheck = true,
			labelsep=space}
		\label{fig:Prev_Discr_Rot_Right}
	\end{minipage}
	\begin{minipage}[t]{0.327\textwidth}
		\includegraphics[width=0.96\textwidth]{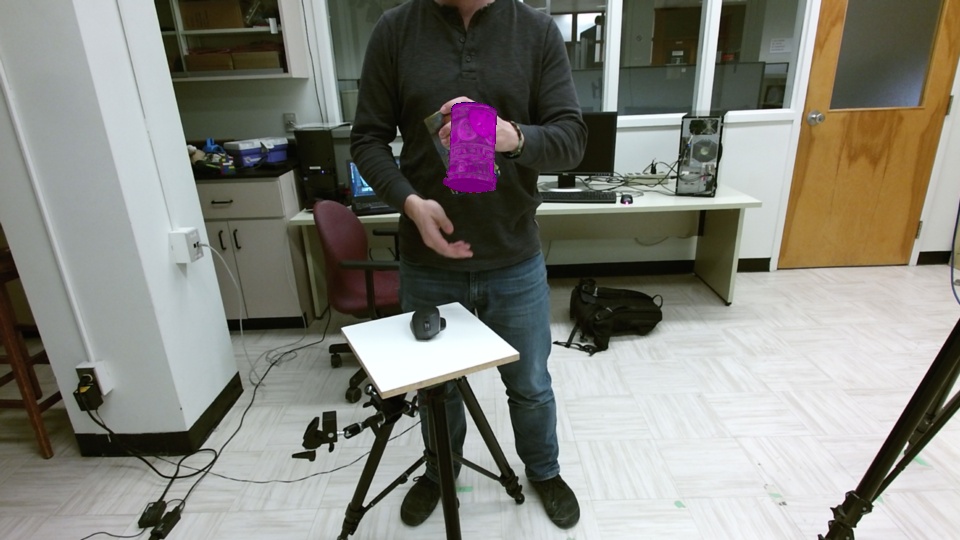}\captionsetup{singlelinecheck = true, 
			labelsep=space}
		\label{fig:Prev_Discr_Rot_Error}
	\end{minipage}
	\begin{minipage}[t]{0.327\textwidth}
		\includegraphics[width=0.96\textwidth]{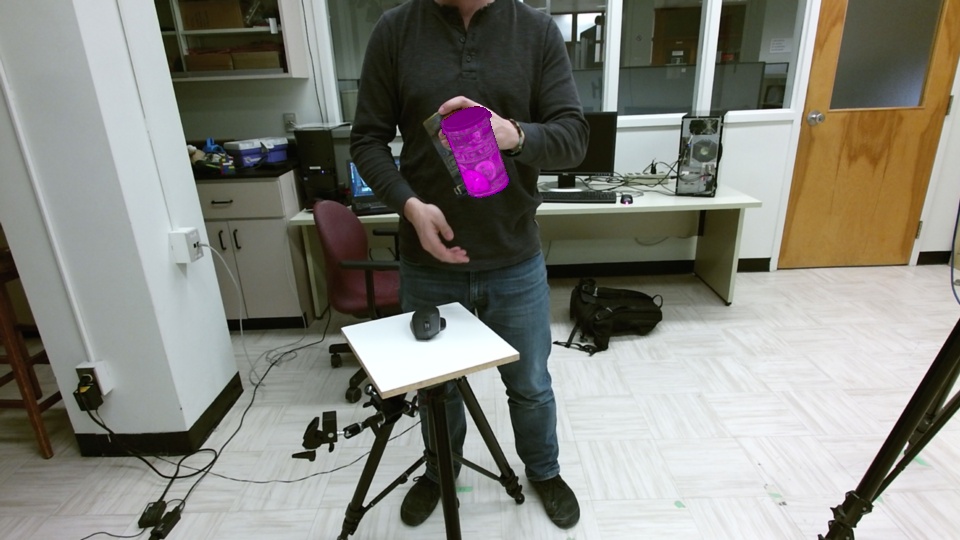}\captionsetup{singlelinecheck = true,
			labelsep=space}
		\label{fig:Discr_Rot_Error}
	\end{minipage}
	\hfill
	\vspace{-0.5cm}
	\caption{\footnotesize{Comparison of the Rotational Estimation for the 127th frame of the "Hard Interaction Scenario" without/with our proposed reflective symmetry handling algorithm. (a) The ground-truth prediction for frame 127. (b) The rotation estimation of the 126th frame. Rotational error between the prediction and ground truth is small. (c) An erroneous rotational estimation of the tracker for frame 127. The prediction is discarded and replaced by the one of the previous timeframe. The error is kept small and (although larger than in the 126th frame) reflective ambiguity is not propagated to the next timestep via the previous pose feedback rendering.}}
	\vspace{-0.6cm}
\end{figure}
\label{fig:reflective}

\medskip
\noindent
\textit{\textbf{Symmetric Object Handling: Continuously Rotational Symmetry}}
In order to form G, we train a batch (of size $B_{2}$) of separate Euler angle triplets  $\mathbf{\hat{g}} \in \R^{3}$. At each timestep, one of them is selected and it is converted to a rotation matrix $\hat{G^{*}}$, which gets right-multiplied with $\Delta \hat{R}$ before being weighted by the parameters of $\Lambda_{(G.S.)}$ (see eq.\ref{eqn:Track_Loss}). Aiming to select the appropriate parameter from the batch, we  train a linear classification layer on top of the first fully connected layer of the tracker. Moreover, we encourage the symmetry triplets to be as uniform as possible by incorporating an appropriate penalty to the overall loss function. To test our approach for the symmetric object case, we used the cylindrical ``Cookie Jar'' model of \cite{Garon_2018}, the shape of which has only one axis of continuous symmetry. Consequently, we estimate a single rotational symmetry parameter, that of the object-centric z-axis (and keep the rest to 0). On the other hand, in the previous case, we define all three of its axes as axes of reflective symmetry. Before the conversion, that parameter is passed through a tanh function and multiplied by $\pi$ to constrain its values.

\vspace{-.5cm}
\subsubsection{Overall Loss}
As a result, our overall tracking loss function is formulated as: 
\vspace{-.2cm}
\begin{equation}
\begin{gathered}\label{eqn:Track_Loss}
\footnotesize L_{Track}(\Delta \hat{\mathbb{P}},\Delta \mathbb{P})=e^{(-v_{1})} \cdot \textit{\footnotesize{MSE}}[(\Delta \hat{\mathbf{t}},\Delta \mathbf{t})]  +v_{1} + v_{2}+ \\ 
+e^{(-v_{2})} \cdot \textit{arcos}\left( \frac{\Tr \left( (\Delta \hat{R} \cdot \hat{G}^{*} \cdot \Lambda_{(G.S.)} )^{T} \cdot (\Delta R \cdot \Lambda_{(G.S.)}) \right)  -1 }{2}\right)
\normalsize
\end{gathered}
\end{equation}\\[-4mm]
Using a similar external multi-task learnable weighting scheme ($\mathbf{s}=[s_{1},s_{2},s_{3}]^{T}$) as in eq.(\ref{eqn:Track_Loss}), we combine our primary learning task, pose tracking, with the two auxiliary ones: clutter and occlusion handling. Both $\mathbf{s}$ and $\mathbf{v}$ are initialized to $\mathbf{0}$.

\vspace{-.2cm}
\begin{equation}\begin{gathered}\label{eqn:MTL_Loss}
Loss = e^{(-s_{1})} \cdot L_{Track} +e^{(-s_{2})} \cdot L_{Unoccl} +  
e^{(-s_{3})} \cdot L_{Foregr} 
 + s_{1} + s_{2} + s_{3} 
\end{gathered}
\end{equation}\\[-4mm]

For objects with continuously rotational symmetry the loss becomes:
\vspace{-.4cm}
\begin{equation}
    Loss^{(Symm)}=Loss+e^{(-s_{4})} \Big( \frac{1}{B}\sum_{b=1}^{B} \frac{1}{\xi_{b}} \Big)+s_{4}, \textnormal{   with   } 
\end{equation}\\[-8mm]
\begin{equation} \xi_{b}=\frac{1}{B_{2}(B_{2}-1)} \sum_{j=1}^{B_{2}} \sum_{k \neq j} d^{(Geod)}_{Rot}(\hat{G}_{k},\hat{G}_{j}),
\end{equation}
The extra term added to the multi-task loss is a penalty that guides the classification layer(s) to select the proper rotational symmetry parameter(s) at each timestep. It encourages the geodesic rotational distances between all pairs of parameters in the batch to be maximum and, thus, ultimately converge to as a uniform distribution as possible. Here, we train $B_{2}$=64 such parameters for each continuous rotational component. 

\vspace{-0.3cm}
\subsection{Data Generation and Augmentation}\label{subsec:data_augmentation}

Following \cite{Garon_2017}, for our network (Fig.\ref{fig:Ours}), we generate two synthetic RGB-D pairs I(t),$\hat{\textnormal{I}}(t)$, but we alter their sampling strategy using the \say{Golden Spiral} approach \cite{Leopardi_2007}, and we modify the augmentation procedure of \cite{Garon_2017,Garon_2018} as follows: Firstly, we blend the object image with a background image, sampled from a subset of the SUN3D dataset \cite{Xiao_2013}. We also mimic the procedure of \cite{Garon_2017,Garon_2018} in rendering a 3D hand model-occluder on the object frame with probability $60\%$. A twist we added, is preparing our network for cases of $100\%$ occlusion, by completely covering the object by the occluder for $15\%$ of the occluded subset. Note that both the foreground and unoccluded object binary masks are kept during both of these augmentation procedures. Hence, we can use them as ground truth segmentation signals for clutter extraction and occlusion handling in our auxiliary losses to supervise the corresponding spatial attention maps. We add to the ``Observed'' frame pair I(t): (i) Gaussian RGB noise, (ii) HSV noise, (iii) blurring (to simulate rapid object movement), (iv) depth downsampling and (v) probabilistic dropout of one of the modalities, all with same parameters as in \cite{Garon_2018}. With a probability of $50 \% $, we change the image contrast, using parameters $\alpha \sim{U(0,3)}$, $\beta\sim{U(-50,50)}$ (where U($\cdot$) is a uniform distribution) and gamma correction $\gamma\sim{U(0,2)}$ with probability $50\%$, to help generalize over cases of illumination differences between rendered and sensor generated images. Instead of modelling the noise added to the ``Observed'' Depth modality with an ad-hoc Gaussian distribution as in \cite{Garon_2018}, we consider the specific properties of Kinect noise \cite{Nguyen_2013} and model it with a 3D Gaussian noise (depending on depth and the ground truth object pose), used for simulating the reality gap between synthetic and real images.
 Its distribution consists of a product of an z-Axial: $n_A \sim {\mathcal{N}(0,\sigma_{A})}$  and two z-Lateral: $n_{L_{x}} \sim {\mathcal{N}(0,\sigma_{L_{x}})}$, $n_{L_{y}} \sim {\mathcal{N}(0,\sigma_{L_{y}})}$  1D distributions that vary with the object depth z and its angle around the y-axis, $\theta_{y}$: with standard deviations $\sigma_{A}$,$\sigma_{L_{x}}$,$\sigma_{L_{y}}$ respectively. The rest of the preprocessing follows \cite{Garon_2017}.

\vspace{-0.3cm}
\section{Evaluation and Results}
\label{sec:evaluation_and_results}

\vspace{-0.2cm}
\subsection{Implementation Details}
\label{subsec:implementation}
 
We use ELU activation functions, a minibatch size of 128, Dropout with probability 0.3, Adam optimizer with decoupled weight decay \cite{Loshchilov_2017} by a factor $1e^{-5}$, learning rate $1e^{-3}$ and a scheduler with warm restarts \cite{Loshchilov_2016} every 10 epochs. All network weights with ELU activation function (except those transferred from ResNet18 \cite{He_2016}) are initialized via a uniform K.He \cite{He_2015} scheme, while for all those with a symmetric one we use a corresponding Xavier \cite{Glorot_2010} distribution. Since the Geodesic distance suffers from multiple local minima \cite{Fletcher_2010}, following \cite{Mahendran_2017}, we first warm-up the weights, aiming first to minimize the LogCosh\cite{Belagiannis_2015} loss function for 25 epochs. Then, we train until convergence, minimizing the loss (\ref{eqn:MTL_Loss}). The average training time is 12 hours in a single GeForce 1080 Ti GPU.

\begin{table}[t]
\centering
\includegraphics[width=.13\linewidth,]{Images/Dragon/Dragon.png} \hfill
\includegraphics[width=.1\linewidth]{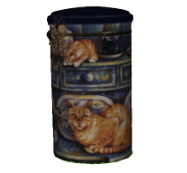} \hfill
\includegraphics[width=.12\linewidth]{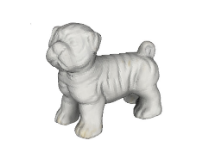}\hfill
\includegraphics[width=.08\linewidth]{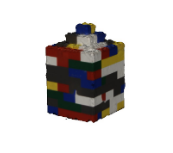}\hfill
\includegraphics[width=.17\linewidth]{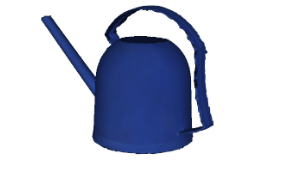}
\\
\centering
\adjustbox{max width=0.9\linewidth}{
\begin{tabular}{c c|c|c|c|c}
\makecell{\centering Object} &\multicolumn{5}{c}{\small{Attributes}}\\
\cline{2-6}
& \small{Size} & \small{Symmetry} & \small{Shape} & \small{Texture} & \small{Distinctive parts}\\
\hline
``Dragon'' & Medium & No & Complex & Rich & Yes\\
``Cookie Jar'' & Medium & Rotoreflective & Simple & Poor and Repetitive & No\\
\say{Dog} & Medium & No & Complex& Almost None & Yes\\
``Lego'' & Small & No & Complex & Rich and Repetitive & No\\
\say{Watering Can} & Big & No & Simple & Poor & Yes\\
\bottomrule
\end{tabular}}
\caption{\footnotesize{Characteristics of the five objects we test our approach on. The fact that there are no two identical items validates the generalization capabilities of our tracker.}}
\label{tab:Objects}
\vspace{-0.8cm}
\end{table}

\vspace{-0.3cm}
\subsection{Dataset and Metrics}\label{subsec:dataset_metrics}

We test our approach on all the \say{Interaction} scenarios and the highest percentage \say{Occlusion} scenarios of \cite{Garon_2018}, which are considered the most challenging. As in \cite{Garon_2018}, we initialize our tracker every 15 frames, and use the same evaluation metrics.

Due to limited computational resources, we produced only 20,000 samples, whose variability covers the pose space adequately enough to verify the validity of our experiments, both for the ablation study and the final experimentation. In the following tables, similarly to \cite{Garon_2018}, we report the mean and standard deviation of our error metrics, as well, as the overall tracking failures (only in the final experimentation section).

\subsection{Ablation Study}
\label{subsec:ablation}
\vspace{-0.2cm}
In this section, we discuss our main design choices and we demonstrate quantitative results that led to their selection.
\vspace{-0.4cm}
\subsubsection{Hierarchy choices for the attention modules}
\label{subsubsec:attention_modules_hierarchy}

Here, we justify the need for both attention modules of our architecture (Fig. \ref{fig:Ours}). We build upon the network proposed by \cite{Garon_2018} and we firstly introduce a single convolutional attention map just for occlusion handling \label{abl:Occl_handl}. Then, we explore the possibility for a seperate attentional weighting of the ``Observed'' feature map for foreground extraction, prior to the occlusion one, \label{abl:hierarch_attentions} and we, finally, leverage both in parallel\label{abl:parallel_attentions} and add their resulting maps altogether.  

\begin{table}[t]
\centering
\adjustbox{max width=0.80\linewidth}{
\begin{tabular}{ c c c }
& \makecell{Translational Error(mm)}& \makecell{Rotational Error(degrees)}\\
\midrule
Garon et al. \cite{Garon_2018} & 34.38 $\pm$ 24.65 & 36.38 $\pm$ 36.31 \\
Only occlusion & 17.60 $\pm$ 10.74 & 37.10 $\pm$ 35.08 \\
Hierarchical clutter $\&$ occlusion& 14.99 $\pm$ \textbf{9.89} & 39.07 $\pm$ 33.22 \\
\textbf{Parallel clutter} $\mathbf{\&}$ \textbf{occlusion} & $\mathbf{14.35} \pm 10.21$ & $\mathbf{34.28 \pm 29.81}$\\
\bottomrule
\end{tabular}}
\caption{\footnotesize{Comparison of different attentional foreground/occlusion handling configurations added to the baseline architecture of Garon et al.\cite{Garon_2018}.}}\label{tab:Tr_Rot_Hierarchical_vs_Parallel}
\vspace{-0.3cm}
\centering
\adjustbox{max width=0.45\linewidth}{
\begin{tabular}{ c c }
&\small{Rotational Error(degrees)}\\
\midrule
\small{Garon et al. \cite{Garon_2018}}& 36.38 $\pm$ 36.31 \\
\small{Rotational MSE}&  46.55 $\pm$ 40.88 \\
\small{Geod.}&  37.69 $\pm$ 35.39 \\
\small{Geod.+\cite{Zhou_2019}}&  14.90 $\pm$ 21.76 \\
\small{Geod.+\cite{Zhou_2019}+$\Lambda_{(G.S.)}$}& $\mathbf{9.99 \pm 13.76}$\\
\bottomrule
\end{tabular}}
\caption{\footnotesize{The evolution of the proposed rotation loss, on the baseline architecture of Garon et al. \cite{Garon_2018} (without our proposed attention modules).}}
\label{tab:Tr_Rot_Rotation_losses_choices}
\vspace{-0.3cm}
\centering
\adjustbox{max width=0.80\linewidth}{
\begin{tabular}{ c c c }
&\small{Translational Error(mm)}&\small{Rotational Error(degrees)}\\
\midrule
\small{Garon et al. \cite{Garon_2018}}& 48.58 $\pm$ 38.23 & 36.38 $\pm$ 36.31 \\
\small{Steady Weights}& 11.83 $\pm$ 8.94  &  10.98 $\pm$ 16.74  \\
\small{Recursive Batch Standarization \cite{Welford_1962}} &  13.97 $\pm$ 10.23 &  14.76  $\pm$  19.24 \\
\small{Learnable Weights \cite{Kendall_2017}} & \textbf{11.63} $\mathbf{\pm}$ \textbf{8.79}  &  \textbf{8.31} $\pm$ \textbf{6.76} \\
\bottomrule
\end{tabular}}
\caption{\footnotesize{Comparison of different multi-task weighting schemes.}}\label{tab:Tr_Rot_MultiTask_Weight}
\vspace{-0.8cm}
\end{table}

The comparison of \textbf{Table \ref{tab:Tr_Rot_Hierarchical_vs_Parallel}} establishes not only the need for both attentional modules in our design, but also that the parallel layout is the optimal one. We can observe the effect of parallel connection in Fig.\ref{subfig:attentions} as both attentions present sharp peaks. We can, also, observe a visual tradeoff between the parallel attentions: while the object is not occluded (either in steady state or when moving), the module responsible for foreground extraction is highlighted more intensely than the occlusion one. As the object gets more and more covered by the user's hand, the focus gradually shifts to the module responsible for occlusion handling. Note that this is not an ability we explicitly train our network to obtain, but rather a side effect of our approach, which fits our intuitive understanding of cognitive visual tracking. Moreover, although our supervising signals are uniform, both attentional modules learn to highlight specific keypoints of interest during the tracker's inference. \newline
\tab We, also, observe another interesting property: the tracker learns to handle self-occlusion patterns as well. For example, for the ``Dragon'' object of Fig.\ref{subfig:attentions}, it learns to ignore the one wing when it is in front and focuses on the other one, at the back, if its appearance is more distinctive of the pose. The same can be said for the legs of the \say{Dog} model of Fig.\ref{subfig:dog}. This is a clue that this module has, indeed, learned  the concept of occlusions and has not overfitted to the shape of the user's hand, the occluder model that was used for training.

\vspace{-0.4cm}
\subsubsection{Contributions of the rotation Loss components}
\label{subsubsec:contributions_rotation_losses}

We demonstrate the value of every component included in our rotation loss (leaving symmetries temporarily out of study), by:  (i) regressing only the rotational parameters with the baseline architecture of \cite{Garon_2018}, (ii), replacing the MSE loss with the Geodesic one, (iii), replacing the rotation parameterization of \cite{Garon_2018} with the continuous one of (\ref{eqn:6D_Cont_Rot}), and, (iv) including the Inertial Tensor weighting of each rotational component.

\begin{table}[t!]
\centering
\adjustbox{max width=0.9\linewidth}{
\begin{tabular}{c c c c}
&\footnotesize{Translational Error(mm)}&\footnotesize{Rotational Error(\textit{degrees})}& \small{$d^{(o)}_{A}(\hat{r}_{z},r^{GT}_{z})$}\\
\midrule
\makecell{\small Garon et al.\cite{Garon_2018} } & 10.75 $\pm$ 6.89 & 23.53 $\pm$ 18.85 & 10.60 $\pm$ 38.90\\
\makecell{\small Ours } \normalsize & 10.87 $\pm$ 8.14 & 20.55 $\pm$ 18.06 & 4.60 $\pm$ 35.42\\
\makecell{\small Ours+Unique,frozen \\
learnable symmetry parameter} \normalsize & 11.38 $\pm$ 8.94  &  10.98 $\pm$ 16.74 & 2.98 $\pm$ 25.07\\ 
\makecell{\small Ours+Regression of \\
learnable symmetry parameter } \normalsize &  13.03 $\pm$ 6.88 &  17.25 $\pm$  12.40 & 2.84 $\pm$ 24.95\\
\makecell{\small Ours+Mean of a batch of frozen,\\
learnable symmetry parameters } \normalsize & 11.98 $\pm$ 9.23 & 13.84  $\pm$ 11.87   & 2.16 $\pm$ 29.25 \\
\textbf{\makecell{\small Ours+Optimal selection out of \\
a batch of frozen, \\
learnable symmetry parameters} } \normalsize & \textbf{10.43} $\mathbf{\pm}$ \textbf{6.63}  &  \textbf{9.57} $\mathbf{\pm}$ \textbf{10.01} &\textbf{2.07} $\pm$ \textbf{24.52}\\
\bottomrule
\end{tabular}}
\caption{\footnotesize{Comparison of different selection methods for the continuously rotational symmetry parameter.}}\label{tab:Symmetry}
\vspace{-0.8cm}
\end{table}

\textbf{Table \ref{tab:Tr_Rot_Rotation_losses_choices}} indicates the value that translation estimation brings to rotation estimation, as when the former's regression is excluded, the latter's performance decreases. Moreover, \textbf{Table \ref{tab:Tr_Rot_Rotation_losses_choices}} justifies our progressive design selections in formulating our rotation loss, as with the addition of each ambiguity modelling, the 3D  rotation error decreases, starting from $46.55^{o} \pm 40.88^{o}$ and reaching $9.99^{o} \pm 13.76^{o}$.

\vspace{-0.5cm}
\subsubsection{Weighting the Multi-Task Loss}\label{subsubsec:multitask_weighting}
Here, we explore various weighting schemes of the multiple loss functions of our approach, both the main and the auxiliary ones. Our first approach is the crude addition of the tracking and the two Binary Cross Entropy losses \label{abl:steady_weight}. 
A second one is standarizing the three losses by subtracting their batchwise means and dividing by their batchwise standard deviations, that we calculate using the Welford algorithm\cite{Welford_1962}. 
Lastly, we consider the learnable weighting strategy that we, ultimately, utilize \label{abl:learnable_weight}. In quantitative comparison of \textbf{Table\ref{tab:Tr_Rot_MultiTask_Weight}} the scheme of \cite{Kendall_2017} emerges as the clear favourite.

\vspace{-0.5cm}
\subsubsection{Comparison of Continuously Rotational Symmetry Regression methods (``Cookie Jar'' model)}

\begin{figure}[t!]
    \centering
    \includegraphics[width=.1\linewidth]{Images/Cookiejar/CookieJar.png}
    \includegraphics[width=.32\linewidth,height=.15\linewidth]{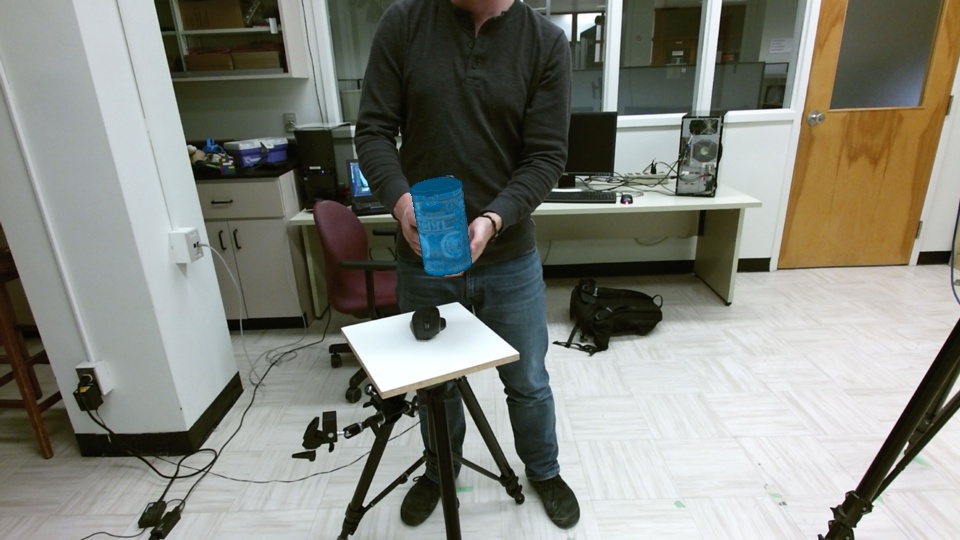}
    \includegraphics[width=.32\linewidth,,height=.15\linewidth]{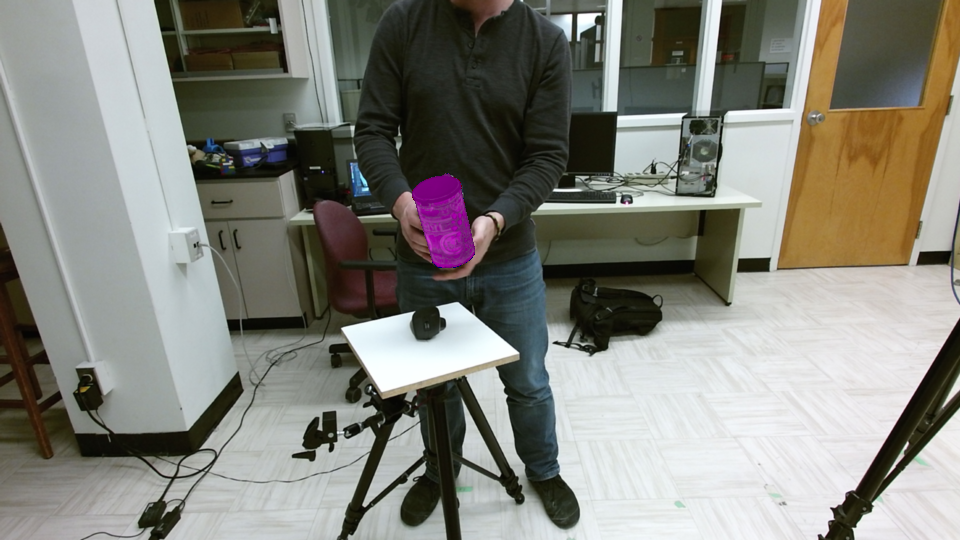} 
    \begin{minipage}[b][0.1\textheight][s]{0.15\textwidth}
  \centering
  \includegraphics[width=.5\textwidth]{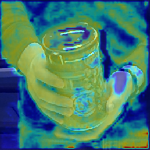}
  \vfill
  \includegraphics[width=.5\textwidth]{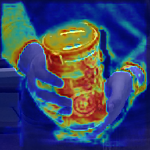}
\end{minipage}
    \\
    \includegraphics[width=.27\linewidth]{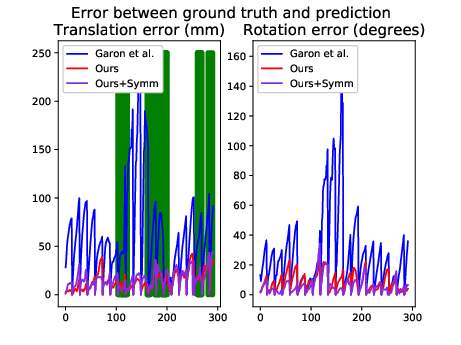}
    \includegraphics[width=.27\linewidth]{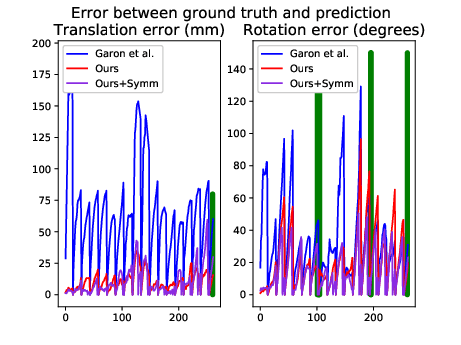}
    \includegraphics[width=.27\linewidth]{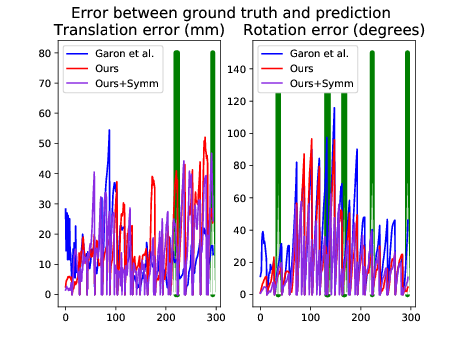}
    \vspace{-0.3cm}
    \caption{\footnotesize Comparison of the SoA\cite{Garon_2018} \textit{(light blue)} and our \textit{(pink)} approaches for the ``Cookie Jar'' in 3 scenarios: ``Translation Only'', ``Rotation Only'' and ``Full Interaction''.} \normalsize \label{subfig:cookie}

    \includegraphics[width=.1\linewidth]{Images/Lego/Lego.png}
    \includegraphics[width=.32\linewidth,height=0.15\textwidth]{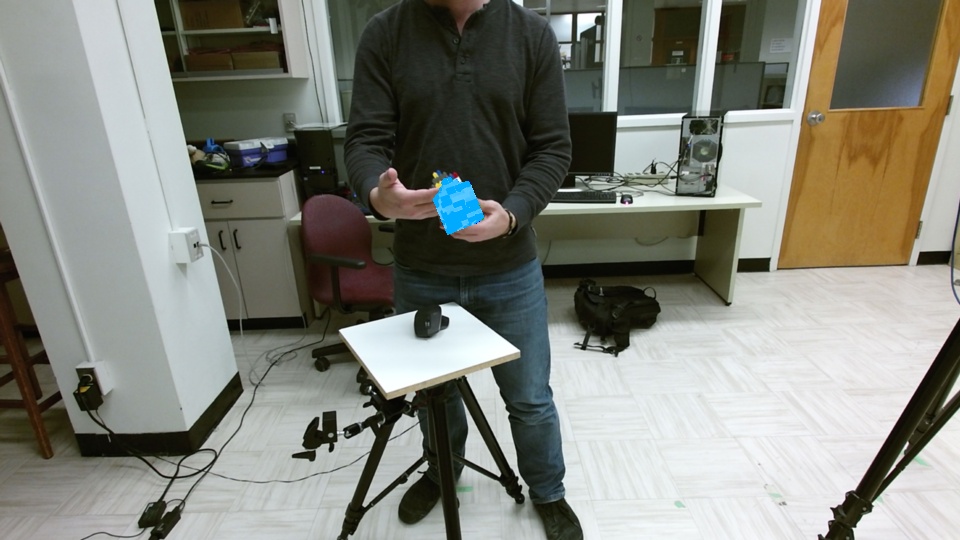} 
    \includegraphics[width=.32\linewidth,height=0.15\textwidth]{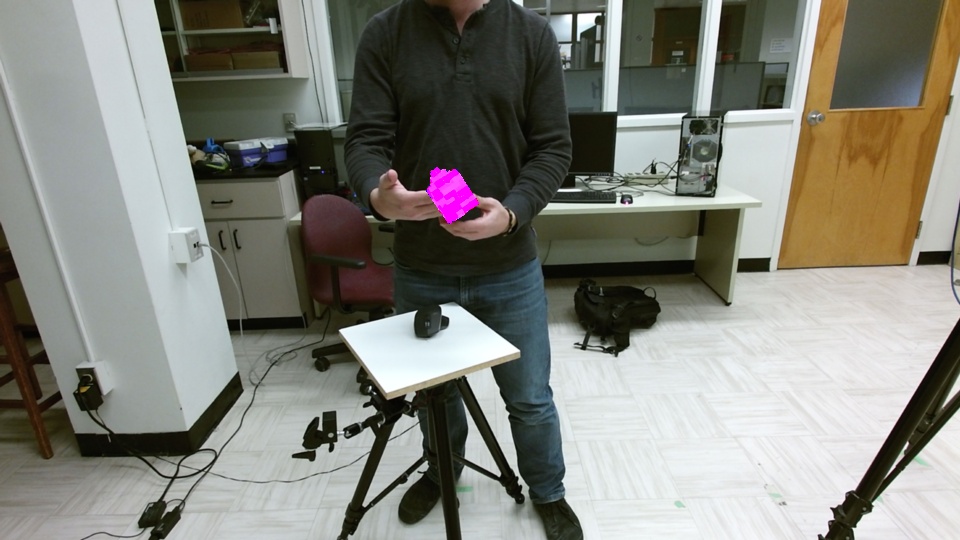} 
    \begin{minipage}[b][0.1\textheight][s]{0.15\textwidth}
  \centering
  \includegraphics[width=.5\textwidth]{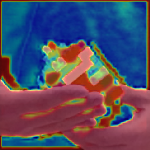}
  \vfill
  \includegraphics[width=.5\textwidth]{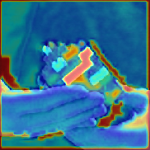}
\end{minipage}
    \\
    \centering
 \includegraphics[width=.27\linewidth]{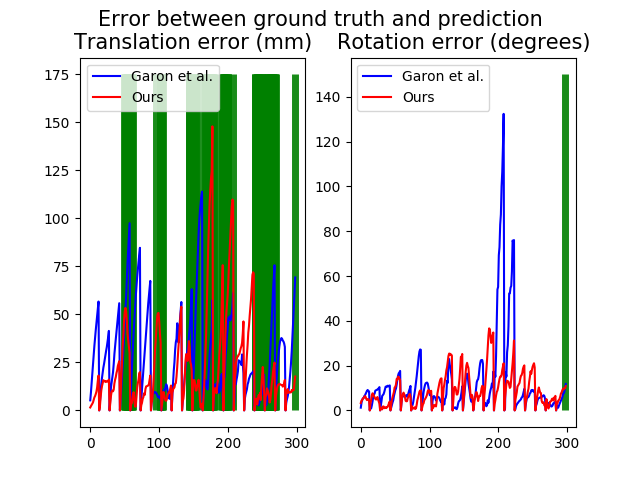}
    \includegraphics[width=.27\linewidth]{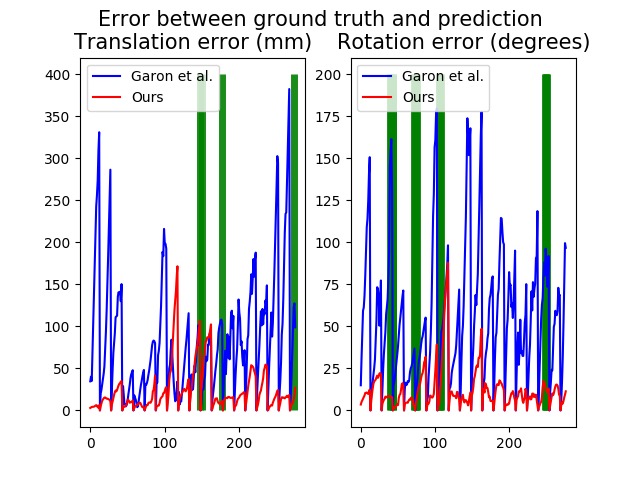}
    \includegraphics[width=.27\linewidth]{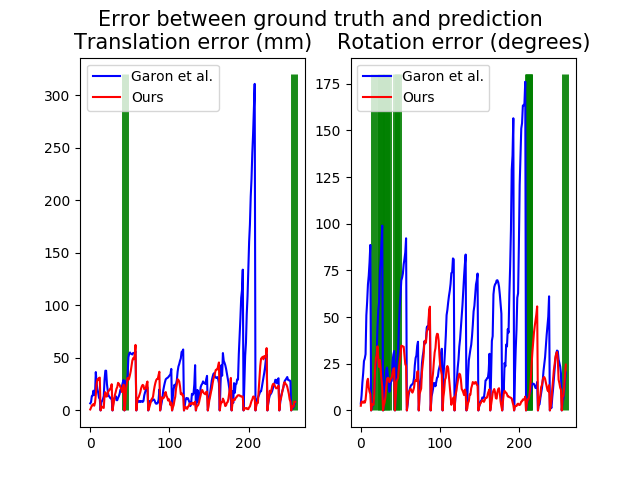} 
\vspace{-0.3cm}
    \caption{\footnotesize{Comparison of the SoA\cite{Garon_2018} \textit{(light blue)} and our \textit{(pink)} approaches for the ``Lego'' in 3 scenarios: ``Translation Only'','Full' and ``Hard Interaction''.} }
\vspace{-0.4cm}
\label{subfig:lego}
\end{figure}

In our effort to disentangle the rotation estimation and the continuously rotational symmetries, we try different configurations for optimally choosing the appropriate parameter(s): (i) learning a unique symmetry parameter over all possible pose changes in the training set and keeping it frozen during inference, (ii) regressing a different one per pose pair, (iii) learning a batch of them and taking their average during inference and, ultimately, (iv) selecting the optimal from the aforementioned batch using an appropriate classification layer while encouraging the values of this batch to be as uniform as possible, at the same time. This freedom of ours resides from the fact that the minimization of the tracking loss w.r.t. the symmetry matrix $\hat{G^{*}}$ (see \cite{Bregier_2017}) does not explicitly impose a global-solution constraint. This time, the comparison is done w.r.t. the full approach of Sect.\ref{sec:methodology} that does not account for symmetries and, apart from the classic 3D translational and rotational errors, we, also, report the Euler angle error for the z-component. As we can see in \textbf{Table \ref{tab:Symmetry}}, our proposed approach is the ``golden medium'' between accuracy and robustness since both its mean and standard deviation are the lowest across all metrics.

\vspace{-0.5cm}
\subsection{Experimental Results}\label{subsec:experimental_results}
\begin{figure}[t!]
    \centering
    \includegraphics[width=.1\linewidth]{Images/Dog/Dog.png}
    \includegraphics[width=.32\linewidth,height=.15\linewidth]{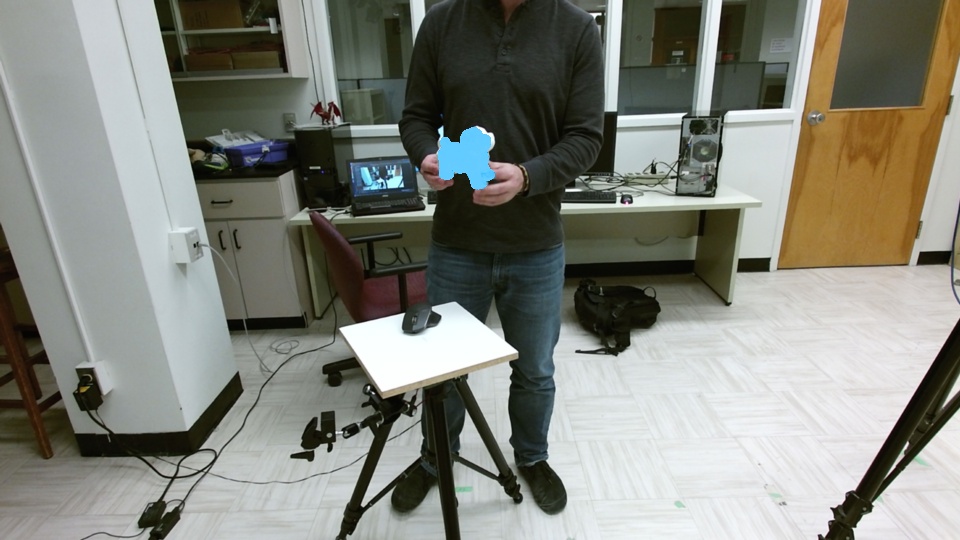}
    \includegraphics[width=.32\linewidth,height=.15\linewidth]{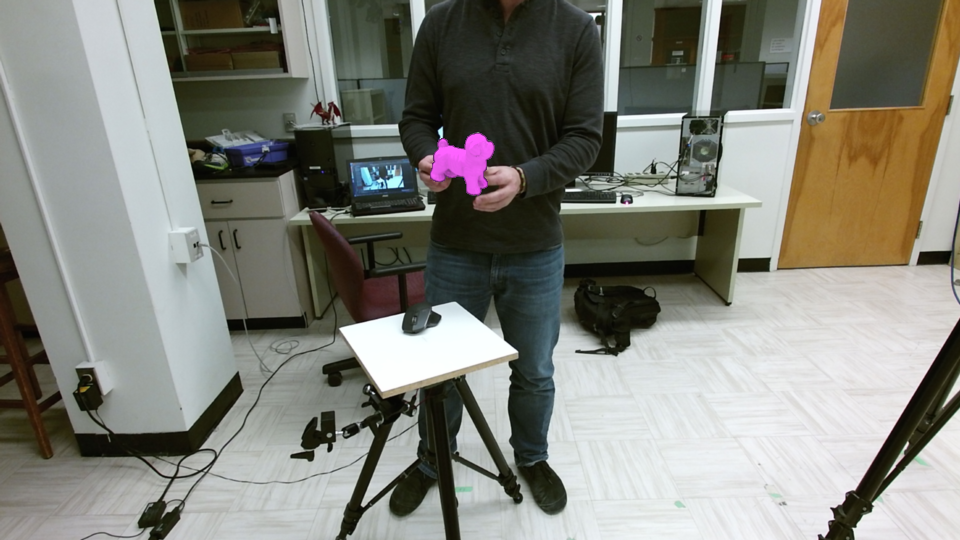}
    \begin{minipage}[b][0.1\textheight][s]{0.15\textwidth}
  \centering
  \includegraphics[width=.5\textwidth]{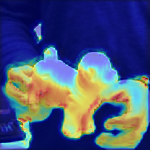}
  \vfill
  \includegraphics[width=.5\textwidth]{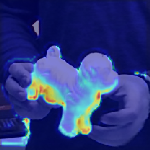}
\end{minipage}
    \\
    \centering
    \includegraphics[width=.27\linewidth]{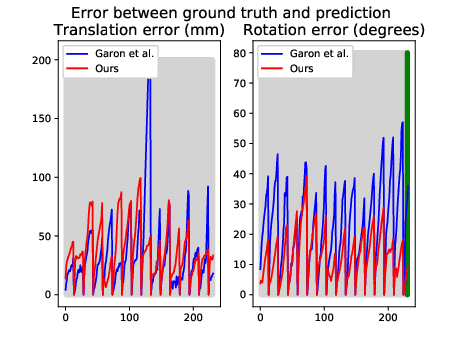}
    \includegraphics[width=.27\linewidth]{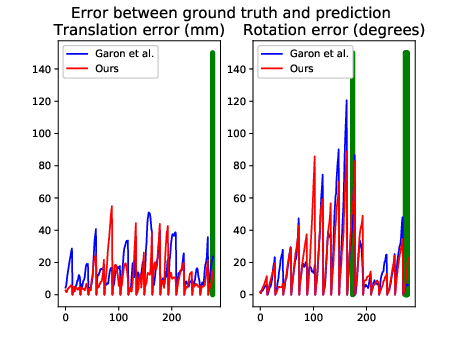}
    \includegraphics[width=.27\linewidth]{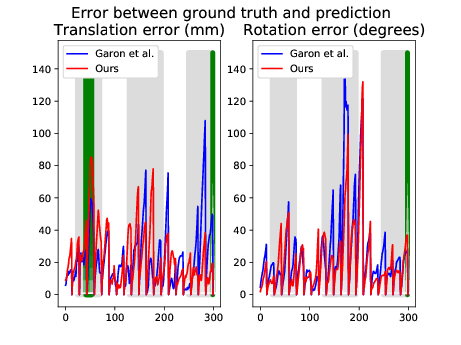}
    \vspace{-0.3cm}
    \caption{\footnotesize{Comparison of the SoA\cite{Garon_2018} \textit{(light blue)} and our \textit{(pink)} approaches for the ``Dog'' in 3 scenarios: ``$75\%$ Vertical Occlusion'', ``Rotation Only'' and ``Hard Interaction''.} }
\label{subfig:dog}
    \centering
    \includegraphics[width=.1\linewidth]{Images/Water/Water.png}
    \includegraphics[width=.32\linewidth,height=0.15\textwidth]{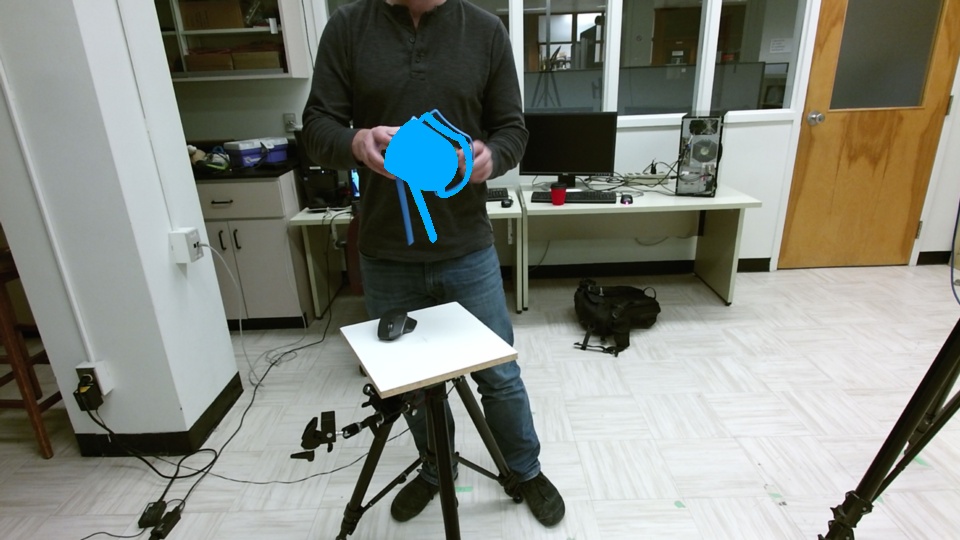} 
    \includegraphics[width=.32\linewidth,height=0.15\textwidth]{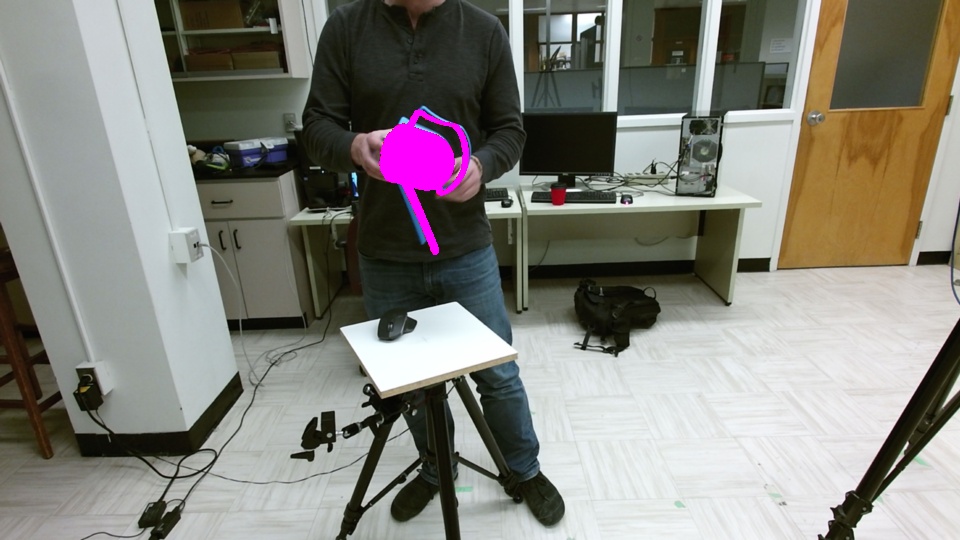} 
    \begin{minipage}[b][0.1\textheight][s]{0.15\textwidth}
  \centering
  \includegraphics[width=.5\textwidth]{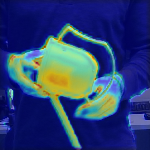}
  \vfill
  \includegraphics[width=.5\textwidth]{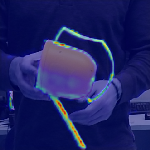}
\end{minipage}
    \\
    \centering
    \includegraphics[width=.27\linewidth]{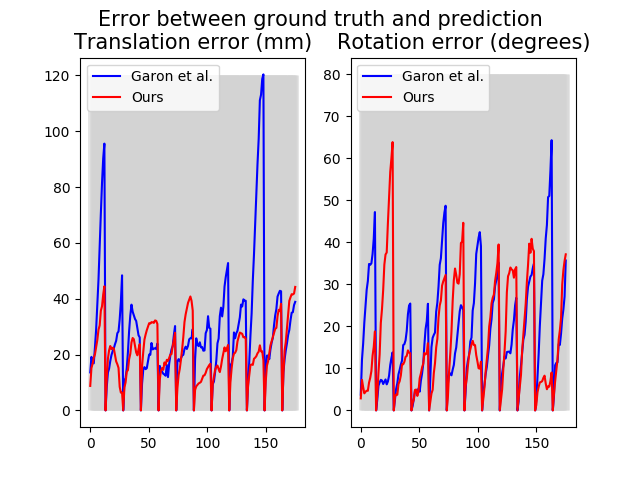}
    \includegraphics[width=.27\linewidth]{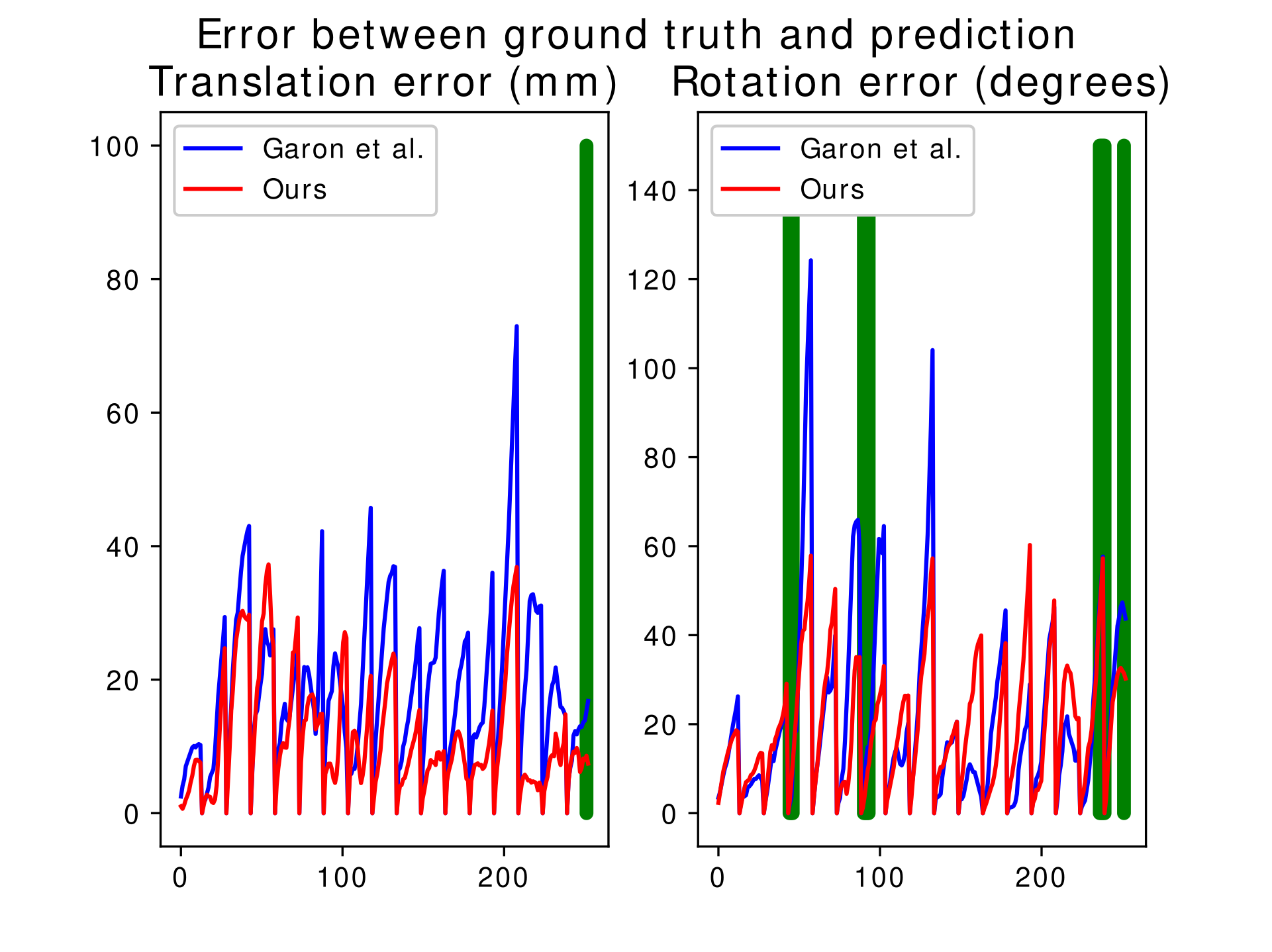}
    \includegraphics[width=.27\linewidth]{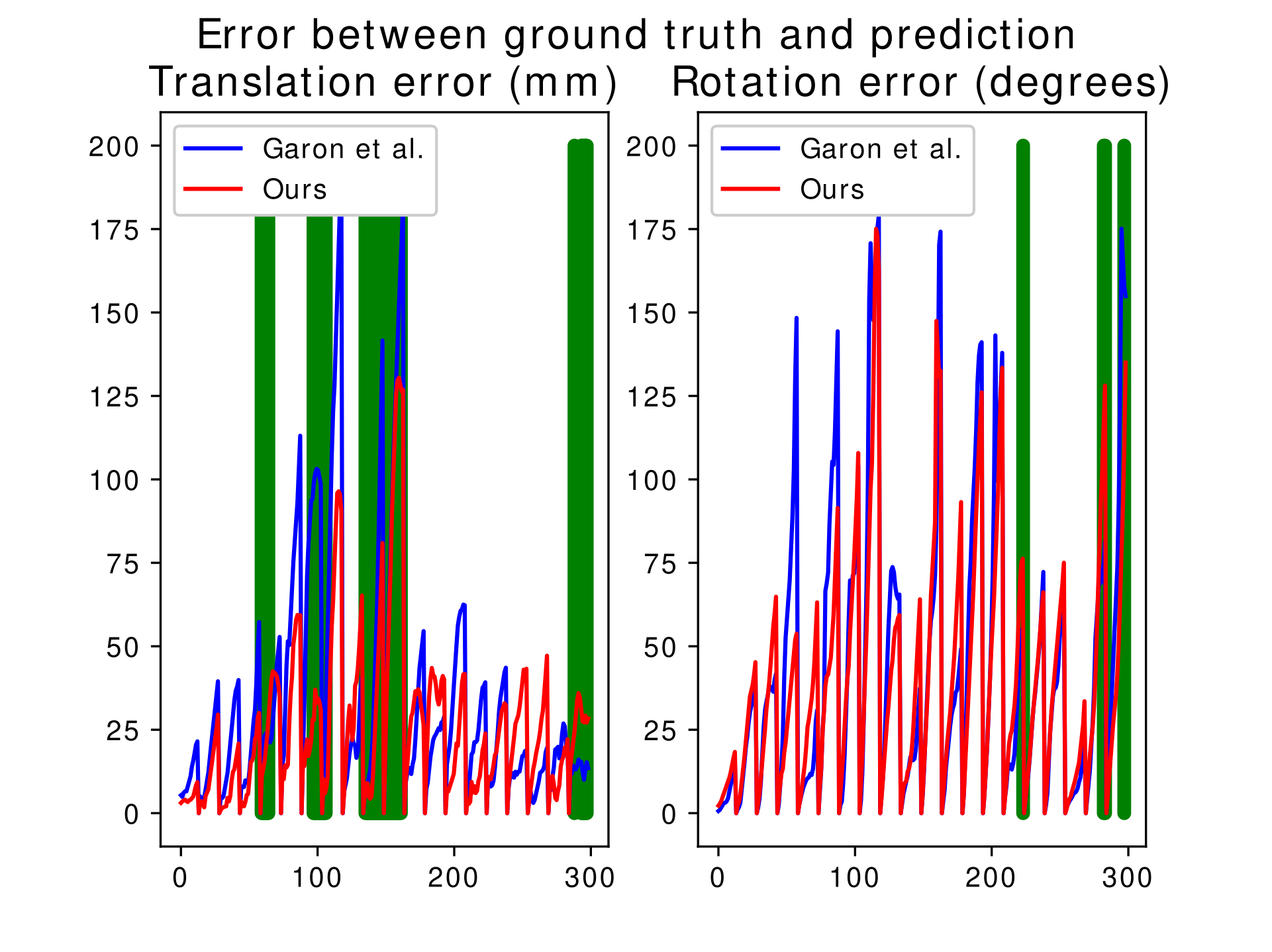}
\vspace{-0.3cm}
    \caption{\footnotesize{Comparison of the SoA\cite{Garon_2018} \textit{(light blue)} and our \textit{(pink)} approaches for the ``Watering Can'' in 3 scenarios:``$75\%$ Horizontal Occlusion'', ``Rotation Only'' and ``Full Interaction''.} }
\label{subfig:water}
\vspace{-0.7cm}
\end{figure}

\vspace{-0.2cm}
According to our ablation study, we proceed to merge our parallel attention modules with the Geodesic rotation loss of eq.(\ref{eqn:Track_Loss}), along with the remaining elements of Sect.\ref{sec:methodology}. We evaluate our method on five objects of dataset \cite{Garon_2018}: the ``Dragon'', the ``Cookie Jar'', the ``Dog'', the ``Lego'' block and the ``Watering Can'', aiming for maximum variability (see \textbf{Table \ref{tab:Objects}}). In Fig.~\ref{subfig:attentions}-\ref{subfig:water}, we plot the 3D translational and rotational errors in three randomly selected scenarios for each object both for the SoA and our tracker.

Evidently, the object most benefited by our methodological improvements is the ``Dragon''. Since its geometry is the most complex, its texture is rich and it has several distinctive parts that stand out of the user's grip, both our geometric modelling and the parallel attention modules find their best application in this case. When the user's hand occludes parts of the ``Dragon'', the attention shifts to its body parts of interest that stand out of the grip, like its neck, wings or tail (Fig.\ref{subfig:attentions}). For the symmetric ``Cookie Jar'', the differences between our method and the baseline are lower. The attentions' effect is less prominent here since this model is of simpler, symmetric shape and poorer texture. This replaces the distinctive clues of the dragon case with ambiguities, denying the corresponding modules of the ability to easily identify the pose. Alongside the ``Lego'' model, they make the most out of the reflective symmetry handling algorithm as they avoid large abrupt errors that propagate to future frames. Although our CNN primarily focuses on the object's shape, appearance seems to play a significant role in its predictions, as well, since the errors of the ``Dog'' and the ``Watering Can'' models, the less textured ones, decrease more mildly. The foreground attention map aids disentangling the ``Dog'' model from its background, a table of the same color, in the ``75$\%$ Occlusion'' scenarios. On the other hand, for the ``Watering Can'', the most ambiguities are presented when viewpoint-induced symmetries appear, with the effects of our modelling declining in this case.    

The accuracy of our approach exceeds that of the SoA \cite{Garon_2018}, across all objects and in almost all scenarios, especially for 3D rotations. According to  \textbf{Table\ref{tab:results}}, both errors are generally lower (in terms of both mean and standard deviation) and our tracker fails equally or less often. It presents aggravated errors in fast object motions more rarely than \cite{Garon_2018} and handles both static and dynamic high-percentage occlusion patterns better. Also, it not only keeps track of the object's 3D position under severe occlusions, but extends this property to 3D rotations, as well. Although more computationally intense than \cite{Garon_2018}, it runs in 40 fps.

\begin{table}[t!]
\centering
\adjustbox{max width=1.0\linewidth}{
\begin{tabular}{p{4.6cm}ccc|ccc}
\makecell{\centering Approach} &\multicolumn{3}{c}{\makecell{\small{75$\%$ Horizontal Occlusion}}} & \multicolumn{3}{c}{\makecell{\small{75$\%$ Vertical Occlusion}}}  \\
\cmidrule(r){2-4}\cmidrule(l){5-7}
& \makecell{\small{Translational Error(\textit{mm})}} & \makecell{\small{Rotational Error(\textit{degrees})}} & \makecell{\small{Fails}} 
& \makecell{\small{Translational Error(\textit{mm})}} & \makecell{\small{Rotational(\textit{degrees})}} & \makecell{\small{Fails}} \\
\hline
Garon et al.\cite{Garon_2018} ('Dragon) & 16.02 $\pm$ \textbf{8.42} & 18.35 $\pm$ 11.71 & 13 & 18.20 $\pm$ 11.81 & 14.66 $\pm$ 12.98 & 13 \\
\textbf{Ours}(``Dragon'') & $\mathbf{12.68}$ $\mathbf{\pm}$ 11.49 & $\mathbf{13.00}$ $\mathbf{\pm}$ $\mathbf{9.14}$ & $\mathbf{10}$ &$\mathbf{ 12.87 \pm 10.49 }$ & $\mathbf{  13.14 \pm 8.85 }$ & \textbf{8}\\
\hline
Garon et al.\cite{Garon_2018}(``Cookie Jar'') & 21.27 $\pm$ 9.74 & 21.90 $\pm$ 13.97& 17& 20.77 $\pm$ \textbf{6.88} & 24.86 $\pm$ 13.64 & 20\\
Ours(``Cookie Jar'') & 9.51 $\pm$ 4.17 & 15.48 $\pm$ 9.50 & 15 & 20.97 $\pm$ 7.32 & 16.14 $\pm$ 10.06  & 15 \\
\textbf{Ours+Symm.}(``Cookie Jar'')& $\mathbf{6.37}$ $\mathbf{\pm}$ $\mathbf{2.14}$ & \textbf{7.22} $\pm$ \textbf{3.97}  &$\mathbf{11}$  & $\mathbf{19.01}$ $\mathbf{\pm}$ 7.53 & \textbf{13.00} $\pm$ \textbf{7.49} & $\mathbf{14}$\\
\hline
Garon et al.\cite{Garon_2018}(\say{Dog}) & 37.96 $\pm$ 23.39 & 47.94 $\pm$ 31.55 & \textbf{21} & \textbf{32.84} $\pm$ 34.07 & 22.44 $\pm$ 13.60 & 21 \\
\textbf{Ours}(\say{Dog}) & \textbf{24.43} $\pm$ \textbf{18.92} & \textbf{17.24} $\pm$ \textbf{12.41} & 25 & 36.53 $\pm$ \textbf{22.39} & \textbf{12.67} $\pm$ \textbf{7.95} & \textbf{20} \\
\hline
Garon et al.\cite{Garon_2018}(``Lego'') & \textbf{68.25} $\pm$ 46.97 & 40.04 $\pm$ 47.37 & 28 & 40.04 $\pm$ 47.37 &35.30 $\mathbf{\pm}$ 31.32 & 20 \\
\textbf{Ours}(``Lego'') & 72.04 $\pm$ \textbf{34.10}  & \textbf{18.41} $\pm$ \textbf{13.84} & \textbf{28} & \textbf{12.92}  $\pm$ \textbf{5.73}  & \textbf{12.92} $\pm$ \textbf{9.02}  & \textbf{20} \\
\hline
Garon et al.\cite{Garon_2018}(\say{Watering Can}) & 21.59 $\pm$ 11.32 & 23.99 $\pm$ \textbf{16.95} & 14 & 32.76 $\pm$ 24.12 & 26.74 $\pm$ 19.05 & 18 \\
\textbf{Ours}(\say{Watering Can}) & \textbf{20.71} $\pm$ \textbf{10.24} & \textbf{17.00} $\pm$ 18.99 & \textbf{13} & \textbf{17.66} $\pm$ \textbf{17.95} & \textbf{13.46} $\pm$ \textbf{10.43} & \textbf{12}\\
\bottomrule
\end{tabular}}

\centering
\adjustbox{max width=1.0\linewidth}{
\begin{tabular}{p{4.6cm}ccc|ccc}
\makecell{\centering Approach} &\multicolumn{3}{c}{\makecell{\small{Translation
Interaction}}} & \multicolumn{3}{c}{\makecell{\small{Rotation
Interaction}}}  \\
\cmidrule(r){2-4}\cmidrule(l){5-7}
& \makecell{\small{Translational Error(\textit{mm})}} & \makecell{\small{Rotational Error(\textit{degrees})}} & \makecell{\small{Fails}} 
& \makecell{\small{Translational Error(\textit{mm})}} & \makecell{\small{Rotational(\textit{degrees})}} & \makecell{\small{Fails}} \\
\hline
Garon et al.\cite{Garon_2018} (``Dragon'') & 41.60 $\pm$ 39.92 & 11.55 $\pm$ 15.58 & 15 &  23.86 $\pm$ 17.44 & 27.21 $\pm$ 22.40 & 15\\
\textbf{Ours}(``Dragon'') & $\mathbf{11.05}$ $\mathbf{\pm}$ $\mathbf{8.20}$ & $\mathbf{3.55}$ $\mathbf{\pm}$ $\mathbf{2.27}$  & $\mathbf{1} $ &$\mathbf{ 9.37 \pm 6.07 }$ & $\mathbf{ 7.86 \pm 6.69 }$ & \textbf{2}\\
\hline
Garon et al.\cite{Garon_2018}(``Cookie Jar'') & 20.43 $\pm$ 25.44 & 17.19 $\pm$ 12.99 & 16 & 10.75 $\pm$ \textbf{5.89} & 23.53 $\pm$ 18.85 &  19\\
Ours(``Cookie Jar'') & 8.64 $\pm$ 8.23 & 8.31 $\pm$ 5.97 & 5 & 10.87 $\pm$ 8.14 & 20.55 $\pm$ 18.06 & 16\\
\textbf{Ours+Symm.}(``Cookie Jar'') & $\mathbf{8.09}$ $\pm$ $\mathbf{7.67}$ & $\mathbf{5.83}$ $\mathbf{\pm}$ $\mathbf{5.50}$ & $\mathbf{3}$  & $\mathbf{9.98}$  $\mathbf{\pm}$ 10.63  & $\mathbf{13.84}$   $\mathbf{\pm}$ $\mathbf{11.87}$ & $\mathbf{16}$ \\
\hline
Garon et al.\cite{Garon_2018}(\say{Dog}) & 58.87 $\pm$ 71.86 & 16.42 $\pm$ 13.51  & 20 & 11.16 $\pm$ 10.28 & \textbf{20.00} $\pm$ 21.31 & 17\\
\textbf{Ours}(\say{Dog}) & $\mathbf{21.64}$ $\mathbf{\pm}$ $\mathbf{22.78}$ & $\mathbf{9.27}$ $\mathbf{\pm}$ $\mathbf{8.03}$ & $\mathbf{14}$ & $\mathbf{10.68}$ $\mathbf{\pm}$ $\mathbf{7.53}$  &$20.07$ $\mathbf{\pm}$ $\mathbf{19.29}$  & $\mathbf{17}$ \\
\hline
Garon et al.\cite{Garon_2018}(``Lego'') & 27.90 $\pm$ \textbf{23.53} & 11.89 $\pm$ 18.50 & 29 & 16.42 $\pm$ 10.90 & 17.83 $\pm$  15.90 & 32 \\
\textbf{Ours}(``Lego'') & $\mathbf{22.66}$ $\mathbf{\pm}$ 24.58 & $\mathbf{9.08}$ $\mathbf{\pm}$ $\mathbf{7.60}$ & $\mathbf{12}$ & $\mathbf{10.13}$ $\mathbf{\pm}$ $\mathbf{6.79}$& $\mathbf{7.22}$ $\mathbf{\pm}$ $\mathbf{4.55}$ & $\mathbf{4}$\\
\hline
Garon et al.\cite{Garon_2018}(\say{Watering Can}) & 24.95 $\pm$ 42.91 & 13.26 $\pm$ 11.34  & 16 & 13.14 $\pm$ $\mathbf{8.99}$ &  22.19 $\pm$ 25.93 & 15\\
\textbf{Ours}(\say{Watering Can}) & $\mathbf{24.30}$ $\mathbf{\pm}$ $\mathbf{21.51}$ & \textbf{8.79} $\pm$ \textbf{6.35} & $\mathbf{16}$ & \textbf{12.22} $\pm$ 9.46 & \textbf{18.66} $\pm$ \textbf{15.51}  & $\mathbf{15}$\\
\bottomrule
\end{tabular}}
\centering
\adjustbox{max width=1.0\linewidth}{
\begin{tabular}{p{4.6cm}ccc|ccc}
\makecell{\centering Apporach} &\multicolumn{3}{c}{\makecell{\small{Full 
Interaction}}} & \multicolumn{3}{c}{\makecell{\small{Hard 
Interaction}}}  \\
\cmidrule(r){2-4}\cmidrule(l){5-7}
& \makecell{\small{Translational Error(\textit{mm})}} & \makecell{\small{Rotational Error(\textit{degrees})}} & \makecell{\small{Fails}} 
& \makecell{\small{Translational Error(\textit{mm})}} & \makecell{\small{Rotational(\textit{degrees})}} & \makecell{\small{Fails}} \\
\hline
Garon et al.\cite{Garon_2018} (``Dragon'') & 35.23 $\pm$ 31.97 & 34.98 $\pm$ 29.46 & 18 & 34.38 $\pm$ 24.65 & 36.38 $\pm$ 36.31 & 17\\
\textbf{Ours}(``Dragon'') &$\mathbf{10.31}$ $\mathbf{\pm}$ $\mathbf{8.66}$  & $\mathbf{6.40}$ $\mathbf{\pm}$ $\mathbf{4.52}$  & $\mathbf{1}$ &$\mathbf{ 11.63 \pm 8.79 }$ & $\mathbf{ 8.31 \pm 6.76 }$ & \textbf{2}\\
\hline
Garon et al.\cite{Garon_2018}(``Cookie Jar'') & \textbf{13.06} $\pm$ \textbf{9.35} & 31.78 $\pm$ 23.78 & 24 & 15.78 $\pm$ 10.43 & 24.29 $\pm$ 20.84 & 15\\
Ours(``Cookie Jar'') & 17.03 $\pm$ 11.94 & 22.24 $\pm$ 20.86 & 21 & 15.29 $\pm$ 16.06 & 16.73 $\pm$ 14.79 & 11 \\
\textbf{Ours+Symm.}(``Cookie Jar'')&  14.63 $\mathbf{\pm}$ 11.19 & $\mathbf{15.71}$ $\mathbf{\pm}$ $\mathbf{13.80}$ & $\mathbf{21}$ & $\mathbf{14.96}$ $\mathbf{\pm}$ $\mathbf{9.06}$ & $\mathbf{15.00}$ $\mathbf{\pm}$  $\mathbf{13.20}$  & $\mathbf{8}$\\
\hline
Garon et al.\cite{Garon_2018}(\say{Dog}) & 37.73 $\pm$ 42.32 & \textbf{20.77} $\pm$ \textbf{19.66} & 23 & 23.95 $\pm$ 38.86 & 24.38 $\pm$ 26.39 & 20\\
\textbf{Ours}(\say{Dog}) &$\mathbf{24.88}$ $\mathbf{\pm}$ $\mathbf{35.85}$  & $28.52$ $\pm$ $25.38$  & $\mathbf{20}$ & $\mathbf{19.32}$ $\mathbf{\pm}$ $\mathbf{15.97}$  &$\mathbf{19.72}$ $\mathbf{\pm}$ $\mathbf{20.17}$  & $\mathbf{19}$ \\
\hline
Garon et al.\cite{Garon_2018}(``Lego'') & 30.96 $\pm$ 31.44 & 22.10 $\pm$ 20.20 & 20 & 30.71 $\pm$ 42.62 & 36.38 $\pm$ 34.99 & 20\\
\textbf{Ours}(``Lego'') & $\mathbf{23.58}$ $\mathbf{\pm}$ $\mathbf{27.73}$ & $\mathbf{11.80}$ $\pm$ $\mathbf{12.28}$ & \textbf{13} & $\mathbf{16.47}$ $\mathbf{\pm}$ $\mathbf{12.95}$ & $\mathbf{14.29}$ $\mathbf{\pm}$ $\mathbf{11.68}$ & $\mathbf{11}$  \\
\hline
Garon et al.\cite{Garon_2018}(\say{Watering Can}) & 33.76 $\pm$ 37.62 & 40.16 $\pm$ 35.90 & 26 & 28.31 $\pm$ 19.49 & 23.04 $\pm$ 24.27 & 28 \\
\textbf{Ours}(\say{Watering Can}) & $\mathbf{19.82}$ $\mathbf{\pm}$ $\mathbf{19.98}$ & $\mathbf{28.76}$ $\mathbf{\pm}$ $\mathbf{30.27}$ & $\mathbf{26}$& $\mathbf{18.03}$ $\mathbf{\pm}$ $\mathbf{14.99}$  &$\mathbf{19.57}$ $\mathbf{\pm}$ $\mathbf{17.47}$ & $\mathbf{23}$ \\
\bottomrule
\end{tabular}}
\caption{\footnotesize{3D Translational and Rotational errors and overall tracking failures in six different scenarios for five employed objects.}}
\label{tab:results}
\vspace{-0.8cm}
\end{table}

\vspace{-0.3cm}
\section{Conclusion}\label{sec:conclusion}
\vspace{-0.2cm}
In this work, we propose a CNN for fast and accurate single object pose tracking. We perform explicitly modular design of clutter and occlusion handling and we account for the geometrical properties of both the pose space and the object model during training. As a result, we reduce both SoA pose errors by an average of $34.03\%$ for translation and 40.01$\%$ for rotation for a variety of objects with different properties. Our tracker exceeds the SoA performance in challenging scenarios with high percentage occlusion patterns and rapid movement and we gain an intuitive understanding of our artificial tracking mechanism.

\vspace{-0.5cm}
\section*{Acknowledgements}\label{sec:acknowledgements}
\vspace{-0.3cm}
This research has been co-financed by the European Union and Greek national funds through the Operational Program Competitiveness, Entrepreneurship and Innovation, under the call RESEARCH CREATE INNOVATE (project code: T1EDK-01248, acronym: i-Walk).

\FloatBarrier
\newpage
\bibliographystyle{splncs04}
\bibliography{strings.bib,main.bib}
\end{document}